\documentclass[preprint,12pt,authoryear]{elsarticle}

\usepackage[colorlinks=true,linkcolor=blue,citecolor=blue,urlcolor=blue]{hyperref}

\usepackage{amssymb}
\usepackage{amsmath}
\usepackage{url}
\usepackage{soul}
\newtheorem{proposition}{Proposition}
\usepackage{tikz}
\usepackage{pgfplots}
\pgfplotsset{compat=1.16}
\usepackage{caption} 
\usepackage{booktabs}
\usepackage{multirow}
\usepackage{colortbl}
\usepackage{xcolor}
\usepackage{adjustbox}

\journal{Neural Networks}
\makeatletter
\def\ps@pprintTitle{%
  \let\@oddhead\@empty
  \let\@evenhead\@empty
  \def\@oddfoot{%
    \footnotesize
    \hfill
    \begin{minipage}{0.75\textwidth}
      \centering
      Published in \textit{Elsevier Neural Networks Journal}, 18 July 2025, Access: \url{https://doi.org/10.1016/j.neunet.2025.107879}
    \end{minipage}%
    \hfill
  }%
  \let\@evenfoot\@oddfoot}
\makeatother

\begin{document}

\begin{frontmatter}


 \author{Tamim Al Mahmud\corref{cor1}}
 \ead{tamimal.mahmud@urv.cat}
\author{Najeeb Jebreel}
 \ead{najeeb.jebreel@urv.cat}
\author{Josep Domingo-Ferrer}
 \ead{josep.domingo@urv.cat}
 \author{David S\'anchez}
 \ead{david.sanchez@urv.cat}
 \cortext[cor1]{Corresponding author}

\title{DP2Unlearning: An Efficient and Guaranteed Unlearning Framework for LLMs}


\address{Universitat Rovira i Virgili,\\
Department of Computer Engineering and Mathematics,\\
CYBERCAT-Center for Cybersecurity Research of Catalonia,\\
Av. Pa\"{\i}sos Catalans 26, 43007 Tarragona, Catalonia}

\begin{abstract}
Large language models (LLMs) have recently revolutionized language processing tasks but have also brought ethical and legal issues. LLMs have a tendency to memorize potentially private or copyrighted information present in the training data, which might then be delivered to end users at inference time. When this happens, a naive solution is to retrain the model from scratch after excluding the undesired data. Although this \emph{guarantees} that the target data have been forgotten, it is also prohibitively expensive for LLMs. Approximate unlearning offers a more efficient alternative, as it consists of \emph{ex post} modifications of the trained model itself to prevent undesirable results, but it lacks forgetting guarantees because it relies solely on empirical evidence. In this work, we present \textit{DP2Unlearning}, a novel LLM unlearning framework that offers formal forgetting guarantees at a significantly lower cost than retraining from scratch on the data to be retained. DP2Unlearning involves training LLMs on textual data protected using $\epsilon$-differential privacy (DP), which later enables efficient unlearning with the guarantees against disclosure associated with the chosen $\epsilon$. Our experiments demonstrate that DP2Unlearning achieves similar model performance post-unlearning, compared to an LLM retraining from scratch on retained data --the gold standard exact unlearning-- but at approximately half the unlearning cost. In addition, with a reasonable computational cost, it outperforms approximate 
unlearning methods at both preserving the utility of the model post-unlearning and effectively forgetting the targeted information.

The code of our experiments is available at\\ \footnotesize\url{https://github.com/tamimalmahmud/LLM-Unlearning/tree/main/DP2Unlearning}.
\end{abstract}



\end{frontmatter}

\section{Introduction}
\label{sec1}

Thanks to training on massive text corpora, large language models (LLMs)
~\citep{achiam2023gpt, team2023gemini, liu2024deepseek} have transformed the landscape of natural language processing (NLP), excelling in various tasks such as question answering~\citep{khashabi-etal-2020-unifiedqa}, translation~\citep{lewis-etal-2020-bart}, and text generation~\citep{lewis-etal-2020-bart}, as well as more complex applications such as education~\citep{malinka2023educational} and recommendation~\citep{10.1145/3627043.3659574}.

Despite their potential, LLMs pose ethical risks~\citep{10.1145/3531146.3533088}. Their ability to memorize data seen during training~\citep{tirumala2022memorization, carlini2023quantifying} can lead to the unintentional generation of {\em private information}~\citep{10179300, carlini2023quantifying} or the reproduction of {\em copyrighted content}~\citep{chang-etal-2023-speak,karamolegkou-etal-2023-copyright}.
For example, \citep{carlini2021extracting} have extracted hundreds of verbatim text sequences from GPT-2 training examples, which contained personally identifiable information (PII) such as names, phone numbers, and email addresses.
In \citep{li-etal-2023-multi-step}, it was shown that despite the measures taken to prevent the generation of sensitive content by OpenAI’s ChatGPT and the ChatGPT enhanced Bing search engine, adversarially designed prompts could still allow PII extraction from these models.
\citep{karamolegkou-etal-2023-copyright} found that LLMs memorize many copyrighted text fragments, including complete descriptions of LeetCode problems. Recently, proprietary algorithms have revealed large-scale verbatim reproduction of copyrighted material by LLMs, including content from NYT articles, works by Ta-Nehisi Coates and Stephen King, academic articles, song lyrics, and business publications~\citep{linkedinLouisHunt}.

Legal frameworks such as the GDPR~\citep{GDPR} in the EU and the CCPA~\citep{CCPA} in the US
have been established to protect privacy and intellectual property in AI systems. 
The GDPR emphasizes the Right to Be Forgotten (RTBF) for prompt deletion of personal data, while copyright laws balance creator rights with fair use in the US~\citep{Chapter155:online} and quotation rights in the EU~\citep{Directiv50:online}. All of this presents a pressing challenge for LLM managers. 

A naive approach to \emph{forget} memorized private and copyright protected information from trained LLMs involves retraining the LLM from scratch after excluding the data to be forgotten. Although this approach provides forgetting guarantees, it is impractical for LLMs because the computational expense of processing each forgetting request is prohibitively high. 

Machine unlearning~\citep{jang-etal-2023-knowledge,yao2023large,maini2024tofu} is emerging as a promising approach to achieve efficient forgetting.  
It refers to the process of selectively forgetting specific knowledge learned by an LLM without affecting unrelated knowledge.
Based on their guarantee of forgetting, the unlearning methods can be categorized into \emph{exact unlearning} and \emph{approximate unlearning}~\citep{10.1145/3603620}. Exact unlearning methods ensure complete forgetting of unwanted data~\citep{bourtoule2021machine, hu2024exact}, but are not practical for LLM due to their significant computational time and storage requirements. 
On the other hand, approximate unlearning provides a more efficient alternative, employing various heuristic techniques to remove unwanted knowledge while maintaining model performance~\citep{liu2022continual, maini2024tofu, yao-etal-2024-machine, rafailov2024direct}. However, these approximate methods lack formal forgetting guarantees and rely on empirical evidence, thus failing to meet the RTBF as stated in applicable legal frameworks.

In this work, we present DP2Unlearning, a novel framework for formal forgetting with guarantees that uses $\epsilon$-differential privacy (DP)~\citep{dwork2006calibrating} on a strategically modified training pipeline to make unlearning \emph{easier}, \emph{cheaper}, and \emph{guaranteed}. Our method involves pre-training LLMs on textual data protected with $\epsilon$-DP, which later enables efficient unlearning of specific data points with the guarantees against disclosure derived from the chosen $\epsilon$ parameter. This approach allows LLMs to learn generalizable patterns from the protected data without capturing sample-specific details, which facilitates efficient unlearning through fine-tuning of the retained data.

We demonstrate by means of extensive experiments that DP2Unlearning achieves a similar forgetting and preservation of performance (model utility) to exact unlearning by retraining from scratch, while reducing unlearning costs by nearly half. We also show that DP2Unlearning performs much better than the existing approximate unlearning methods in both preserving the utility of the model post-unlearning and effectively forgetting the targeted data. 

The remainder of this paper is organized as follows. Section~\ref{sec2}
discusses related work on unlearning in LLMs. Section~\ref{sec3} provides background on DP. Section~\ref{sec4} presents our DP2Unlearning framework. Section~\ref{sec5} describes
the experimental setup. Section~\ref{sec6} reports the experimental results and provides extensive comparisons with the baseline methods. Conclusions and future directions are collected in Section~\ref{sec7}.
The appendices provide additional experimental results.

\section{Related works}
\label{sec2}

We briefly review the literature on exact and approximate unlearning.

\subsection{Exact unlearning} 
Exact unlearning methods, while offering guarantees for complete data removal, are computationally expensive. The simplest approach is to retrain from scratch on the data to be retained; however, this becomes impractical for LLMs due to the high costs involved. Some works focus on more efficient methods
for exact unlearning.

\textbf{SISA} \citep{bourtoule2021machine} is a generic exact unlearning framework that can be applied to a variety of ML tasks and models, including LLMs. SISA makes retraining less expensive by sharding training data and slicing shards. Model training checkpoints are recorded after each slice, allowing more efficient retraining from the checkpoints corresponding to the slices containing the forget data. Although it provides exact forgetting guarantees, SISA is not practical for LLMs due to the high computational and memory costs associated with 
model saving, checking points, retraining, and inference~\citep{Blanco-Justicia2025}. Also, while increasing the number of shards reduces unlearning costs, 
it raises training and inference costs --a different model
is trained for each shard, and inference
involves an ensemble decision based on the outputs
of all shard-level models-- and
decreases the preservation of the ensemble-level model performance due to 
increased heterogeneity of the shard-level models.

\textbf{Adapter Partition and Aggregation (APA)} is an exact unlearning technique for the LLM recommendation system (LLMRec) that preserves the inference speed intact~\citep{hu2024exact}. It works by dividing the training data into disjoint shards and retraining only the adapters that contain the information that will be forgotten. APA reduces the cost of retraining, but suffers from poor generalizability, insufficient scalability for large data sets, and high memory consumption.

\subsection{Approximate unlearning} 
\label{subsec2.2}

Approximate unlearning aims to adjust a trained model to eliminate specific knowledge without the need for retraining. Although these methods may not ensure complete formal forgetting, they offer a more computationally feasible alternative. In the following, we review popular approximate unlearning methods that will also serve as baselines for comparison in our experiments.

\textbf{Gradient ascent (GA)} adjusts the parameters of LLM to increase the loss associated with specific data, making it less probable for the model to retain and reproduce that information. Unlike traditional retraining on data to be retained, which focuses on minimizing loss to enhance learning, GA tends to suppress unwanted information by maximizing the loss associated with it.
Very recently, several researchers \citep{liu2024rethinking, maini2024tofu, yao-etal-2024-machine} have adopted GA for approximate unlearning in LLM. Although GA is an efficient alternative to retraining, challenges such as catastrophic forgetting --a situation in which the model unintentionally loses important shared knowledge from the retained set while attempting to forget unwanted data from the forget set-- require hybrid approaches that balance forgetting and retention.

\textbf{Gradient Difference (GD)} is a hybrid approach that takes advantage of both gradient ascent and gradient descent \citep{liu2022continual}. Unlike gradient ascent, which only increases the error in unwanted information, gradient difference reduces the difference between the error in the data we want to retain and the error in the data we want to forget. At the same time, it keeps the model performing well on the data we want to keep. Recently, several researchers \citep{maini2024tofu, lev2024faster, trippa2024tau, yao-etal-2024-machine} have applied GD for LLM unlearning. While GD's structured optimization makes it a promising alternative to retraining, optimizing its trade-off between forgetting and model performance remains an ongoing challenge.

\textbf{Kullback-Leibler (KL)} minimization aims to minimize the Kullback-Leibler divergence \citep{hershey2007approximating} between the probability distributions of the model predictions in the retained data before and after unlearning. By maintaining the output distributions, KL-based unlearning ensures that retained knowledge stays constant while unwanted information is gradually suppressed. Due to its efficiency in probabilistic alignment, KL divergence minimization has recently been used for LLM unlearning \citep{yao-etal-2024-machine, wu-etal-2025-rethinking, maini2024tofu}. However, although it provides a less resource-intensive option compared to retraining, it needs additional adjustments to avoid inadvertent loss of knowledge.

\textbf{Preference Optimization (PO)} draws inspiration from direct preference optimization~\citep{rafailov2024direct}. The idea is to modify the model so that it refrains from generating unwanted information. Alternative answers are generated for the questions that refer to the data to be forgotten ({\em e.g.}, ``I do not know the answer''). Then, the modified model is fine-tuned to minimize a sum of the loss of these alternative answers for the data to be forgotten and the loss of the correct answers for the data to be retained. Recent work has applied PO~\citep{tian-etal-2024-forget, maini2024tofu} to adjust LLM outputs for desired behavior after unlearning specific information or patterns. Although effective in some cases, PO faces challenges in balancing the two losses considered.



\section{Background on differential privacy}
\label{sec3} 

Differential privacy (DP)~\citep{dwork2006calibrating} is a privacy model that ensures that the outputs \( \mathcal{A}(\mathcal{D}) \) and \( \mathcal{A}(\mathcal{D}') \) of a mechanism $\mathcal{A}$ calculated on two data sets, \(\mathcal{D}\) and \(\mathcal{D}'\), which differ by only one individual's record, remain statistically indistinguishable
up to an exponential factor of a parameter $\epsilon$. The formal requirement to achieve pure \(\epsilon\)-DP is expressed as
\[
P[\mathcal{A}(\mathcal{D}) \in \mathcal{R}] \leq e^{\epsilon} \ P[\mathcal{A}(\mathcal{D}') \in \mathcal{R}]
\]
In this inequality, \(\mathcal{R}\) is a subset of possible output responses that satisfies \(\epsilon\)-DP and \(\epsilon\) is called privacy budget, which controls the level of disclosure protection. 

The inventors of DP suggest that, for meaningful privacy guarantees against disclosure, the privacy budget ($\epsilon$) should not exceed 1~\citep{Dwork_Kohli_Mulligan_2019}; and that values of $\epsilon$ larger than $10$ are too weak to provide effective disclosure protection. 

The original definition of \(\epsilon\)-DP has been extended to ($\epsilon, \delta$)-DP by including an additive term probability of privacy failure (\(\delta\)), where $\delta<1/|\mathcal{D}|$. 
To achieve indistinguishability, DP typically adds calibrated noise to its output.

DP has some interesting properties:
\begin{enumerate}
    \item \label{immunity} {\em Immunity to post-processing:}  
    If a mechanism \(\mathcal{A}\) satisfies \(\epsilon\)-DP or \((\epsilon, \delta)\)-DP, then any post-processing function \(g(\cdot)\) applied to its output also satisfies \(\epsilon\)-DP or \((\epsilon, \delta)\)-DP, respectively.
    \item {\em Sequential composition:} If a mechanism \(\mathcal{A}_1\) satisfies \(\epsilon_1\)-DP, resp. \((\epsilon_1, \delta_1)\)-DP, and the mechanism \(\mathcal{A}_2\) satisfies \(\epsilon_2\)-DP, resp. \((\epsilon_2, \delta_2)\)-DP, then their combined application $\mathcal{A}_{\text{sequential}}$ on the same data set or on non-disjoint data sets satisfies $(\epsilon_1 + \epsilon_2)\text{-DP}$, resp. $(\epsilon_1 + \epsilon_2, \delta_1 + \delta_2)\text{-DP}$.
    \item {\em Parallel composition:} If mechanisms \(\mathcal{A}_1\) and \(\mathcal{A}_2\) both satisfy \(\epsilon\)-DP, resp. \((\epsilon, \delta)\)-DP, and operate on disjoint data sets \(\mathcal{D}_1\) and \(\mathcal{D}_2\), then their combined mechanism 
    $\mathcal{A}_{\text{parallel}}$ satisfies \(\epsilon\)-DP, resp.  $(\epsilon, \delta)\text{-DP}$.
\end{enumerate}

\subsection{DP and disclosure protection in LLMs}
\label{subsec3.1}

DP was originally designed to protect queries to structured databases \citep{dwork2006calibrating}. However, DP can also be used to prevent disclosure in language models. DP-MLM (Differentially Private Text Rewriting Using Masked Language Models) and DP-SGD (Differentially Private Stochastic Gradient Descent)  are two key mechanisms that we leverage in our work.

\subsubsection{DP-MLM}
\label{DP-MLM}
DP-MLM \citep{meisenbacher-etal-2024-dp} enforces DP on the textual training data. In a privacy-oriented context, DP-MLM should be applied to the noun phrases of each of the documents in the training data set, as they are the most informative units of text --without which it is not possible to disclose specific facts (\emph{e.g.}, private information) about the subjects of the data \citep{ppdp-c-sanitized}--.


DP-MLM can be enforced by applying the exponential mechanism, which probabilistically substitutes disclosive terms with semantically similar alternatives while still retaining the general structure of the document. A utility function \(u(w, w')\) evaluates the semantic similarity between the original term \(w\) and a possible replacement \(w'\), which can be calculated based on contextual embeddings and cosine similarity. Specifically, the probability \(P(w'|w)\) of swapping \(w\) for \(w'\) is represented as:
\[P(w' | w) = \frac{\exp(u(w, w')\epsilon)}{\sum_{w'' \in V} \exp(u(w, w'')\epsilon)}
\]
where $V$ denotes the vocabulary (the candidate group of words used for substitution).

\subsubsection{DP-SGD}
\label{DP-SGD}
DP-SGD \citep{10.1145/2976749.2978318} and its variant \citep{kerrigan-etal-2020-differentially} are an optimization algorithm that enforces DP during model training by modifying the conventional stochastic gradient descent (SGD) through \emph{gradient clipping}, \emph{Gaussian noise injection}, and \emph{gradient update}. These processes ensure that each data point contributes in a limited and randomized way and prevents the model from learning disclosure information. 

\begin{enumerate}
    \item \emph{Gradient clipping} restricts the impact of a single data item on model training. The gradient update to a specific parameter \( \nabla \theta \) is clipped using a set value called the clipping norm $C$, which is the maximum allowed value (threshold). The clipped gradient is calculated as
\[
\bar{\nabla} \theta = \nabla \theta \cdot \min\left(1, \frac{C}{\|\nabla \theta\|_2} \right),
\]
where \(\|\nabla \theta\|_2\) represents the $L_2-$norm of the gradient. This process guarantees that no individual data point excessively affects the optimization procedure, thereby safeguarding against information disclosure.

\item \emph{Gaussian noise injection} consists of 
adding Gaussian noise \( \mathcal{N}_r(0, \sigma^2)\) once the clipping is complete. Noise is added to the combined mini-batch gradient to further obfuscate the impact of specific data pieces before executing the update. The noisy gradient update \( \tilde{\nabla} \theta \) that adheres to DP standards is described as 
\[
\tilde{\nabla} \theta = \sum_{j=1}^{bs} \bar{\nabla} \theta_j + \mathcal{N}_r(0, \sigma^2 I),
\]
where \( bs \) is the mini-batch size and \( \sigma^2 \) is the variance of Gaussian noise, adjusted as \(\sigma^2 = \frac{C^2 \log(1.25 / \delta)}{\epsilon^2}\). This ensures that the algorithm satisfies DP protection with privacy  budget \(\epsilon\). 

\item \emph{Gradient update} is the final process that uses the noisy gradient \(\tilde{\nabla} \theta\), the learning rate \( \eta \), and the model parameter at the \(t\)-th iteration \( \theta_t \). Thus, the resulting gradient update rule is \(\theta_{t+1} = \theta_t - \eta \tilde{\nabla} \theta\). This update ensures that no single data point substantially impacts the training of the model. 
\end{enumerate}



\section{DP2Unlearning}
\label{sec4}

The exact unlearning methods aim to completely eliminate the data that must be forgotten. However, in order to comply with privacy and copyright laws, complete data removal is overkill. For example, the GDPR states that it is sufficient to make personal data non-personal (for example, through anonymization) to be beyond the regulation's scope. Similarly, for copyright protection, it is enough to prevent verbatim reproduction of the original source while still preserving the underlying semantics.

This means that, in practice, what we need is \emph{selective but guaranteed} unlearning. That is, the model should exactly forget specific details while still being able to retain the general meaning. Our hypothesis is that privacy models such as \emph{differential privacy}~\citep{dwork2006calibrating}, \emph{k-anonymity}~\citep{Sweeney2002557}, or their variants, can be used to enforce on the trained model outputs this selective or partial removal of training data with guarantees against (detailed) information disclosure.

Given the heterogeneity and lack of structure of the textual documents employed to train LLMs, DP seems the best-suited model for this task, as it allows for enforcing disclosure protection on documents individually and independently. This is particularly beneficial when scaling to large data sets involved in training LLMs. 

On the other hand, $\epsilon$-DP offers \emph{ex ante} guarantees against disclosure, which ensure that no specific data point can be distinguished from other points based on the model output up to an exponential factor depending on $\epsilon$. In practice, this guarantees that the model 
is guaranteed up to that exponential factor not to reproduce any private or copyright-protected information on which DP has been applied. The actual degree of protection against disclosure can be controlled by the privacy parameter \(\epsilon\), which in our case dictates the level of forgetting. For example, larger values of \(\epsilon\) (\emph{e.g.}, \(\epsilon>10\)) offer mild forgetting, which would tend to approximate forgetting, while \(\epsilon=0\) is equivalent to complete exact forgetting. Middle-ground values would probably be the best suited to provide guaranteed but utility-preserving forgetting. 

Using the intuitions above, we propose DP2Unlearning, an LLM construction framework that uses DP with a modified training pipeline to make unlearning \emph{cheaper} and \emph{guaranteed}. The framework is designed in such a way that it can handle forgetting requests efficiently while adhering to privacy and copyright regulations.

\subsection{Method description}
\label{subsec4.1}

\begin{figure}[ht]
    \centering    \includegraphics[width=1\textwidth]{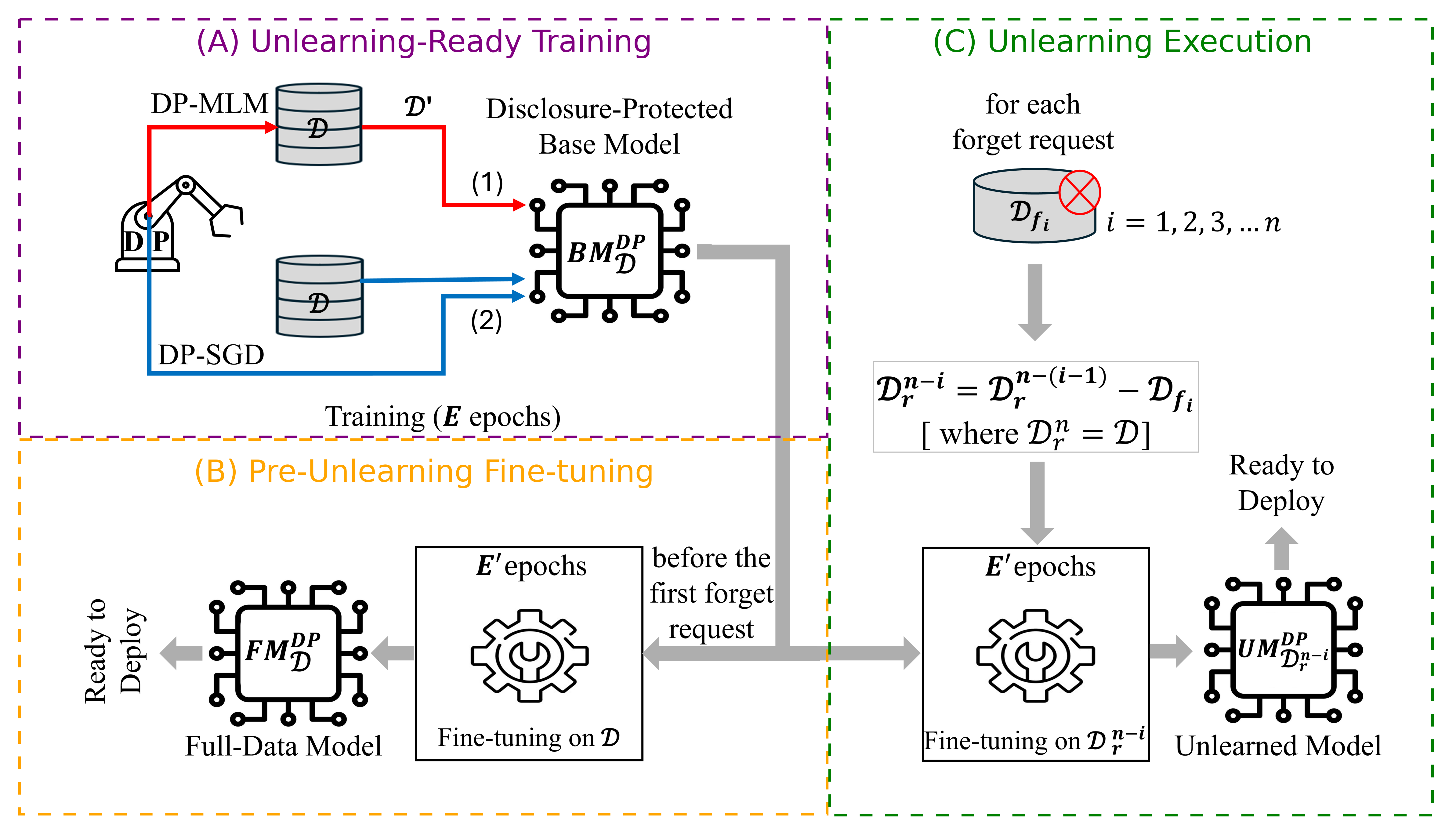}
\caption{Workflow of the proposed DP2Unlearning framework}
    \label{figure1}
\end{figure}

DP2Unlearning operates in three stages: (A) Unlearning-ready training, (B) Pre-unlearning fine-tuning, and (C) Unlearning execution. The framework executes the first two stages, (A) and (B), only once, while the final stage, (C), repeats for each unlearning request. The workflow of DP2Unlearning is depicted in Figure \ref{figure1}.

Stage (A) --\textbf{unlearning-ready training}-- involves training a disclosure-protected base model (or simply a base model), ensuring safeguards against the disclosure of any information that may need to be unlearned. As anticipated in Section \ref{sec3}, this can be accomplished in two different ways:  
\begin{enumerate}
    \item DP-MLM-based data protection: The model is trained on a data set ($\mathcal{D'}$) where the specific text components are obfuscated using DP-MLM. Protection occurs at the individual data point level by substituting disclosure terms, mostly noun phrases, with semantically similar alternatives in a probabilistic manner through the use of an exponential mechanism. The choice of $\epsilon$ directly governs the disclosure guarantee, with smaller values ensuring stronger protection. However, depending on the trade-off between disclosure protection and model utility, a relaxed version ($\epsilon, \delta$)-DP can be implemented, introducing limited uncertainty in token selection. Since DP-MLM functions at the document level, the loss of privacy accumulates across multiple token substitutions, and the total protection adheres to the sequential composition theorem.
    
    \item DP-SGD-based training: DP-SGD directly imposes $(\epsilon,\delta)$-DP constraints during training by adding Gaussian noise to gradient updates, ensuring that individual contributions remain indistinguishable. Although DP-SGD uses $\epsilon$ to control the level of disclosure protection, it also requires a small value $\delta$, since pure $\epsilon$-DP is theoretically impossible with Gaussian noise (with Gaussian noise, there is always a small chance that some updates may exceed the pure $\epsilon$-DP limit, which requires $\delta$ to be taken into account). 
\end{enumerate}

In both cases, the model is trained until convergence for $E$ epochs, resulting in the base model (\(BM_{\mathcal{D}}^{\text{DP}}\)), which we must preserve indefinitely because unlearning rests on it. 
The development of the base model is resource intensive, although it is performed only once. In this respect, an advantage of DP-MLM over DP-SGD is that in the former DP protection does not need to be applied to all training data:
public domain training data sources (\emph{e.g.}, Wikipedia), for which forgetting requests will never apply, can be kept and used `as is', significantly reducing training costs. In contrast, DP-SGD treats all data equally. 



In stage (B) --\textbf{pre-unlearning fine-tuning}-- one tries
to make up for the decrease in model performance that is likely to have occurred as a result of DP-protection in stage (A): the DP-protected base model is expected to exhibit lower model performance compared to a model trained without DP. We hypothesize that by fine-tuning this safeguarded model on the original raw data (\(\mathcal{D}\)), we can restore most (if not all) of its performance; also, due to the incremental nature of fine-tuning and the fact that it is done on the same data sources used for training (even though, in this case, unprotected), it will require significantly less computational resources than retraining the base model on \(\mathcal{D}\) from scratch. This should be possible because initial DP-induced training, while not producing an accurate model, allows it to gain a ``general'' understanding of \(\mathcal{D}\), thus facilitating faster learning of the details and specifics of \(\mathcal{D}\) through fine-tuning. As a result, and as long as there is no forget request, we release a version of $BM_{\mathcal{D}}^{\text{DP}}$ fine-tuned on the complete raw data (\(\mathcal{D}\)) for $E'$ epochs, which corresponds to the initial full-data model (\(FM_{\mathcal{D}}^{\text{DP}}\)) ready for deployment. 
Note that this fine-tuning step uses raw unprotected data to restore the utility lost due to DP protection and, therefore, does not provide DP guarantees. The resulting model is deployed only until the first unlearning request is received. Upon such a request, the system reverts to the DP-protected base model and fine-tunes it on the retain set to offer disclosure protection guarantees on the data to be forgotten.

Stage (C) --\textbf{unlearning execution}-- is run whenever a forget request $\mathcal{D}_{f_i}$ arrives, where $i=1, 2, 3\dots,n$. Specifically, the current deployed model is discarded and the saved base model $BM_{\mathcal{D}}^{\text{DP}}$ is resumed. The latter model is then fine-tuned again using only the data set to be retained, which is
\[
\mathcal{D}_{r}^{n-i} = \mathcal{D}_{r}^{n-(i-1)} - \mathcal{D}_{f_i}, \quad \text{where} \quad \mathcal{D}_{r}^{n} = \mathcal{D}.
\]

This results in an unlearned model ($UM_{\mathcal{D}_{r}^{n-i}}$) with an expected performance similar to retraining from scratch on the retained data, but with \(\epsilon\)-DP or \((\epsilon, \delta)\)-DP guarantees against the disclosure of the data to be forgotten ($\mathcal{D}_{f_i}$). In other words, we achieve DP disclosure guarantees on data to be forgotten because the original data points in \(\mathcal{D}_{f_i}\) have not been seen by the unlearned fine-tuned model (they have
only been seen by the base model under DP protection):
\[ UM_{\mathcal{D}_{r}^{n-i}} = \text{Fine-tuning}(BM_{\mathcal{D}}^{\text{DP}}, \mathcal{D}_{r}^{n-i}, E').\]


\subsection{Unlearning guarantees and computational cost}
\label{subsec4.2}
We prove that either \(\epsilon\)-DP or (\(\epsilon, \delta\))-DP guarantees against disclosure are satisfied for a forgetting request. 

\begin{proposition}[Disclosure]
\label{prop1}
Unlearning with stage (C) fulfills the \(\epsilon\)-DP requirement when utilizing DP-MLM and the (\(\epsilon, \delta\))-DP requirement when utilizing DP-SGD.
\end{proposition}

\textbf{Proof:} Based on the post-processing immunity characteristic of DP (refer to DP characteristics in Section~\ref{immunity}), the \(\epsilon\)-DP or the (\(\epsilon, \delta\))-DP guarantee extend to \( UM_{\mathcal{D}_r^{n-1}}^{\text{DP}} \), ensuring that \(\mathcal{D}_{f_i}\) remains protected against disclosure proportionally to the chosen \(\epsilon\). \hfill $\Box$

\vspace{1ex}
We now analyze the computational cost of the above three stages:



\begin{itemize}
    \item Stages (A) and (B) together {\em incur a large computational cost, but only once}. As mentioned earlier, stage (A) is resource intensive, primarily due to the added computational burden of DP protection. For DP-SGD, the associated cost is higher than that of DP-MLM due to the
    following key factors: (i) per-sample gradient computation and clipping, which adds a computational overhead during training by limiting gradient values to avoid substantial updates that may compromise disclosure information, and (ii) noise injection, which introduces randomness and delays convergence. Consequently, DP-SGD typically incurs longer convergence times than DP-MLM, which uses probabilistic term substitution to obfuscate disclosure information. Moreover, DP-SGD needs to process all data during training, whereas DP-MLM can be used on a sensitive subset (\emph{e.g.}, private or copyrighted data). 
    Stage (B) also introduces a one-time overhead, as it involves fine-tuning the base model with unprotected data to recover model performance for deployment; this is a process that is not required in standard LLM training.
    \item Stage (C) fine-tunes an already trained model (the base model) rather than starting from scratch. This fine-tuning benefits from the knowledge embedded in the base model. Thus, fewer optimization steps are required to recover the performance lost due to DP protection. Therefore, processing a forgetting/unlearning request with stage (C) requires significantly less computational cost than retraining from scratch. Specifically, stage (C) reduces the cost by a factor of $E'/E$, where $E$ is the number of epochs 
    required to train the model from scratch on the data to be retained, and $E'$ is the number of epochs needed to fine-tune the protected base model on the data to be retained.
\end{itemize}
 

\section{Experimental setup}
\label{sec5}
\subsection{Data sets and models}
\label{subsec5.1}
Due to the high computational training costs of LLMs, we cannot afford to train a full-fledged LLM from scratch. Instead, we use pre-trained models --{\em Phi-1.5B} (from Microsoft) and {\em Llama2-7B} (from Meta)-- of varying sizes and capabilities. To have control over the data to be unlearned, we further train these models using additional data sets specifically tailored for evaluating unlearning. 

As additional training data, we used the TOFU \citep{maini2024tofu} data set, a recent benchmark data set specifically designed to evaluate unlearning in LLM. The data set includes 4,000 question-answer pairs derived from 200 varied synthetic author profiles, each comprising 20 question-answer pairs. The data set is synthetically created and intentionally modified to ensure that it does not overlap with the training data typically used to build an LLM. This intentional design makes the TOFU data set a versatile resource for a controlled and unbiased evaluation of unlearning methodologies. 

The TOFU data set is divided into Forget and Retain sets, with varying proportions of 1\%-99\%, 5\%-95\%, and 10\%-90\%, allowing investigation of the impact of different unlearning ratios. Additionally, TOFU includes two supplementary real-world data sets: Real Authors (containing real author-related questions) and Real-World Facts (covering general knowledge questions).

Since TOFU data were intentionally created to
ensure that they were not used to pre-train existing LLMs, 
further training a pre-trained LLM on TOFU allows
experimenting with unlearning, because
training the pre-trained
LLM on the retain subset of TOFU can be viewed as retraining an LLM from scratch on the retain subset. 
We name this setting \emph{retraining from scratch on Retain set (RFS-R)}, which corresponds to \emph{exact unlearning}.
\subsection{Baseline methods}
\label{subsec5.2}
To compare our approach with those of related work, we reproduced several approximate unlearning methods. Specifically, we use the Gradient Ascent (GA), Gradient Difference (GD), Kullback-Leibler Minimization (KL) and Preference Optimization (PO) methods introduced in Subsection~\ref{subsec2.2}. Each method adopts a different strategy:
\begin{itemize}
    \item GA reverses the model updates by maximizing the loss function, thus compelling the model to unlearn previously acquired (unwanted) knowledge.
    \item GD modifies gradients to reduce model retention of the information intended to be forgotten while preserving its usefulness.
    \item KL forces the model output to match a reference distribution by minimizing the KL divergence.
    \item PO utilizes direct preference optimization to modify the model's predictions away from data considered to be forgotten.
\end{itemize}

We implemented these baselines using the TOFU unlearning implementation, which is publicly available at \url{https://github.com/locuslab/tofu}.

\subsection{Evaluation metrics}
\label{evalmetric}

Our evaluation primarily focuses on model utility (\emph{i.e.}, the model's ability to retain useful knowledge post-unlearning) and forget quality (\emph{i.e.}, the extent to which the model effectively forgets unwanted knowledge). More specifically, model utility refers to the model's ability to provide correct responses for information it is supposed to retain, ensuring that unlearning does not harm the model's overall performance.
On the other hand, the forget quality indicates how effectively the model stops giving accurate responses to the information it is supposed to forget.
Note that model utility can also be used to evaluate
forget quality: high post-unlearning utility with respect to the data to be forgotten can be viewed as an indication of poor forget quality.

To evaluate model utility, we used the evaluation metrics --ROUGE-L, conditional probability, and truth ratio-- used in TOFU \citep{maini2024tofu}. 

\begin{itemize}
    \item {\textit{ROUGE}} estimates the similarity between the responses generated by the model and the ground truth answers \citep{lin-2004-rouge}, allowing for minor differences in wording. ROUGE-L evaluates similarity by computing the longest common subsequences (LCS) of words between the model-generated response and the correct (ground truth) answer. This metric gives a score that shows how accurate the content is; even if the words are not exactly the same, it provides a high score for semantically equivalent content. A typical way to calculate ROUGE-L recall is
    \[
    \text{ROUGE-L} = \frac{\text{LCS}(GTT, MGT)}{|GTT|},
    \]
    where $GTT$ is the ground truth text (reference text), $MGT$ is the model generated text (hypothesis), $\text{LCS}(GTT, MGT)$ represents the length of the longest common subsequence between $GTT$ and $MGT$, and $|GTT|$ is the total number of tokens in the reference text.
   
    \item {\textit{Conditional probability}} measures the confidence of the model in its predictions. It helps in assessing retention effectiveness. The formula used for a query and response pair (\( Q, r\)) in the Forget Set and the Retain Set is
    \[
    P(r|Q)^{\frac{1}{|r|}},
    \]
where $P(r|Q)$ is the probability that the model returns the response $r$ when asked query $Q$, normalized by the length $|r|$ of the response
(as done in \citep{cho2014properties}). However, in the Real Authors and Real World Facts data sets, the conditional probability is computed for multiple choice questions as \[\frac{P(r_1|Q)}{\sum_{i=1}^{x} P(r_i|Q)},\] where $r_1$ represents the correct answer among the $x$ options. This process makes it easier to compare answers of different lengths.

\item {\textit{Truth ratio}} (TR) measures how well a model prioritizes a correct response over multiple incorrect responses~\citep{lin-etal-2022-truthfulqa}, reflecting its retained knowledge. Mathematically, the TR score is calculated by dividing the average likelihood of intentionally modified incorrect responses (that is, responses following the same linguistic pattern as the right answer but including inaccuracies that sound believable, yet are incorrect) by the likelihood of a correct paraphrased response (\emph{i.e.}, a semantically accurate rewording of the original answer, making sure that the different wording does not change the meaning). Therefore, this metric quantifies the extent to which the model, even after unlearning the unwanted knowledge, continues to prioritize providing correct responses rather than incorrect ones: \[
    TR = \frac{\frac{1}{|R_{\text{inaccurate}}|} \sum_{\hat{r} \in R_{\text{inaccurate}}} P(\hat{r}|Q)^{\frac{1}{|\hat{r}|}}}{P(\tilde{r}|Q)^{\frac{1}{|\tilde{r}|}}},
    \] 
where $R_{\text{inaccurate}}$ is the set of intentionally modified responses designed to be incorrect, $\tilde{r}$ is the accurate paraphrased response, and $\hat{r}$ is an intentionally modified incorrect response.
\end{itemize}

\textbf{Overall model utility:} To measure the utility of the model as a whole, we evaluated the above three metrics in three data sets: Retain Set, Real-World Facts, and Real Authors. We normalized each of the three evaluation metrics to fall within the range $[0,1]$, where higher values indicate improved retention (utility preservation). To consolidate these metrics into a single model utility, we calculated the harmonic mean of the nine metric values (three values from each of the three data sets). Since the harmonic mean is sensitive to low values, a significantly low score of any of the nine evaluation metrics will disproportionately lower the overall model utility score. 

\textbf{Overall Forget Quality:} To measure forget quality as a whole, we assessed the above three metrics on the forget data set. Due to the intricate nature of LLMs, the evaluation of forgetting quality often relies on statistical tests and changes in data distribution \citep{goel2022towards}. Specifically, we leverage the Kolmogorov-Smirnov (KS) test, a non-parametric statistical test to assess the variations in truth ratios between the unlearned model and a model trained solely on retained data (RFS-R) to check how well a model can forget information. Basically, the KS test evaluates two cumulative distribution functions (CDFs) and computes (i) the KS statistic ($D_{KS}$): the maximum absolute difference between the two CDFs; and (ii) the $p$-value, that is, the probability that the two samples come from the same distribution, which {\em indicates the forget quality}. To find how different the two CDFs are, the KS statistic ($D_{KS}$) can be calculated as
\[
D_{KS} = \max |C_{\text{u}}(x) - C_{\text{r}}(x)|,
\]
where $C_{\text{u}}(x)$ and $C_{\text{r}}(x)$ are the CDFs of truth ratios for unlearned and retain-only models (RFS-R, the benchmark for comparison), respectively. A higher value $D_{KS}$ implies that the CDFs are statistically different and therefore indicates unsuccessful unlearning. On the other hand, a lower value $D_{KS}$ implies that CDFs are statistically identical and therefore indicate successful unlearning. We used a $0.05$ threshold for the probability $p$-value of the $KS$ test: a value below $0.05$ clearly allows us to reject the null hypothesis that the two CDFs are statistically the same, indicating ineffective forgetting; while a high $p$-value ($p \geq 0.05$) indicates that the unlearned model closely follows the retain-only model, indicating good forgetting. The threshold $0.05$ for the $p$-value of the KS test is a widely accepted standard in the testing of statistical hypotheses~\citep{Aslam2019, maini2024tofu}. However, its justification depends on the context of unlearning. Although smaller $p$ values ({\em e.g.}, $p<0.01$) are sometimes preferred in high-sensitivity applications to reduce false positives, in practical unlearning scenarios, a threshold of $0.05$ remains a standard choice. Given our data set size of 4,000 instances, this threshold is justified, as it avoids excessive sensitivity to minor distributional shifts while still detecting meaningful differences in forget quality.

\subsection{Training settings}
We next detail the technical configuration and training parameters used in our experiments to ensure reproducibility and facilitate independent validation of our results. We cover the training settings for unlearning-ready training, pre-unlearning fine-tuning, unlearning execution, DP configurations, and baseline methods unlearning. The code of our experiments is available at\\ \url{https://github.com/tamimalmahmud/DP2unlearning/}

\textbf{Unlearning-ready training:} To determine the optimal number of training epochs for the base model, we analyzed model convergence without DP training. This established a reference point for selecting DP-aware training epochs.

We empirically found that both the \textit{Phi} and \textit{Llama2} models reached a near-perfect ROUGE-L score of $\approx 1.0$ when they were trained on TOFU full data without DP for the $E=10$ and $E=6$ epochs, respectively, as shown in Figure \ref{figure2} and Table \ref{table1}.       
\begin{figure}[ht!]
\centering
\begin{minipage}{0.47\textwidth} 
    \centering
    \begin{tikzpicture}
        \begin{axis}[
            width=0.81\textwidth,
            height=4.5cm,
            xlabel={},
            ylabel={ROUGE-L Score},
            xmin=5, xmax=10,
            ymin=0.6, ymax=1.05,
            xtick={5,6,7,8,9,10},
            ytick={0.60,0.70,0.80,0.90,1.0},
            grid=major,
            legend pos=south east,
        ]
        
        \addplot[
            mark=star, 
            color=blue,
        ]
        coordinates {
            (5, 0.6604)
            (6, 0.7719)
            (7, 0.8945)
            (8, 0.9832)
            (9, 0.9948)
            (10, 1.00)
        };
        \addlegendentry{Phi}
        
        \addplot[
            mark=star, 
            color=red,
        ]
        coordinates {
            (5, 0.9746)
            (6, 0.9980)
        };
        \addlegendentry{Llama2}
        \end{axis}
    \end{tikzpicture}
    \vspace{-0.2cm}    
    \caption{ROUGE-L at different epochs}
    \label{figure2}
\end{minipage}\hfill
\begin{minipage}{0.53\textwidth} 
    \vspace{-1cm} 
    \centering
    \scriptsize 
    \setlength{\tabcolsep}{4pt} 
    \captionof{table}{ROUGE-L at different stages. For {\em Phi}, $E=10$ and $E'=5$. For {\em Llama2}, $E=6$ and $E'=3$.}
    \begin{tabular}{|c|c|c|c|c|}
        \hline
        Model & Pre & Trained & Our & Our \\
              & Trained & without & \( BM_{\mathcal{D}}^{\text{DP}} \) & \( FM_{\mathcal{D}}^{\text{DP}} \) \\
              && DP &  & \\ 
              && $E$ epochs & $E$ epochs & $E'$ epochs\\
        \hline
        Phi & 0.4494 & 1.00 & 0.4233 & 0.9957 \\
        \hline
        Llama2 & 0.3549 & 0.9972 & 0.3834 & 0.9789 \\
        \hline
    \end{tabular}
    \label{table1}
\end{minipage}
\end{figure}

This observation guided our choice: we aligned the DP-aware training epochs with the non-DP training convergence points. We set $E=10$ for {\em Phi} (trained with DP-MLM and DP-SGD) and $E=6$ for {\em Llama2} (trained only with DP-MLM, as DP-SGD is computationally prohibitive for our setup).

\textbf{Pre-unlearning fine-tuning:} 
We empirically determined that fine-tuning for the $E'\approx E/2\approx5$ epochs (for {\em Phi}) and the $E'\approx E/2\approx 3$ epochs (for {\em Llama2}) significantly improved their ROUGE-L scores, bringing them close to $1.00$ (see Table \ref{table1}). 

\textbf{Unlearning execution:} Since unlearning is performed using the same fine-tuning approach as in pre-unlearning fine-tuning, we adopt the same convergence point. Therefore, to process each forget request, we fine-tune the corresponding base model (\( BM_{\mathcal{D}}^{\text{DP}} \)) exclusively on the retain data for $E'\approx E/2\approx 5$ epochs (for \textit{Phi}) and $E'\approx E/2\approx 3$ (for \textit{Llama2}), respectively.

\textbf{Privacy settings for DP-MLM and DP-SGD:}
We experimented with different values of the privacy budget (\(\epsilon= 0.5, 1, 10, 25, 100\)) to find the best balance among disclosure protection guarantees, model performance and computational overhead. Additionally, we optimized the configurations for each DP mechanism as follows: 
\begin{itemize}
    \item for DP-MLM, we set the logit clipping bounds to clip\_min = -5.2093 and clip\_max = 20.3048 to ensure controlled sensitivity when selecting substitute tokens, which we found to provide the best trade-off between disclosure protection and semantic coherence.
    
    \item for DP-SGD, we set the minimum possible $\delta$ to make it nearly equivalent to pure $\epsilon$-DP, as DP-SGD inherently requires \(\delta>0\) as discussed in the methodology. Since achieving pure $\epsilon$-DP ($\delta = 0$) is theoretically impossible, we followed the best practice of setting ($\delta$ to be smaller than $1/|\mathcal{D}|$, see Section~\ref{sec3} for justification), and ensured \(\delta \ll 2.5 \times 10^{-4}\) given the 4,000 instances of the data set. 
\end{itemize}

\textbf{Baseline methods:} To ensure a fair comparison that aligns with our unlearning execution settings, we applied the same fine-tuning settings across all baseline methods. We set $E' = 5$ epochs for Phi and $E' = 3$ epochs for Llama2.

\textbf{Hardware setup:}  
All experiments were performed on an NVIDIA H100 GPU with 80 GB of HBM3 memory. We used a learning rate \(5 \times 10^{-5}\), a weight decay $0.01$, and an effective batch size $16$ (batch size $4$, gradient accumulation steps $4$).  



\section{Experimental results}
\label{sec6}

We now proceed to evaluate the two variants of our DP2Unlearning framework (i) \textbf{DP2U-SGD}, when the model is trained with DP-SGD and (ii) \textbf{DP2U-MLM} when the model is trained on data protected by DP-MLM.

We first analyze the trade-off between disclosure protection and model performance. This is crucial since an increased privacy budget can compromise disclosure protection and thus unlearning guarantees, while a reduced budget degrades the model utility, thereby requiring more fine-tuning effort to recover utility. To identify the best balance, we systematically evaluated key performance metrics at various values of $\epsilon$.

Subsequently, we performed a comparative analysis to evaluate the effectiveness of our approaches by comparing them with exact unlearning through RFS-R and several approximate unlearning baselines (discussed in Section~\ref{subsec2.2}).


\subsection{Balancing disclosure protection and model performance} 
\label{best_epsilon}
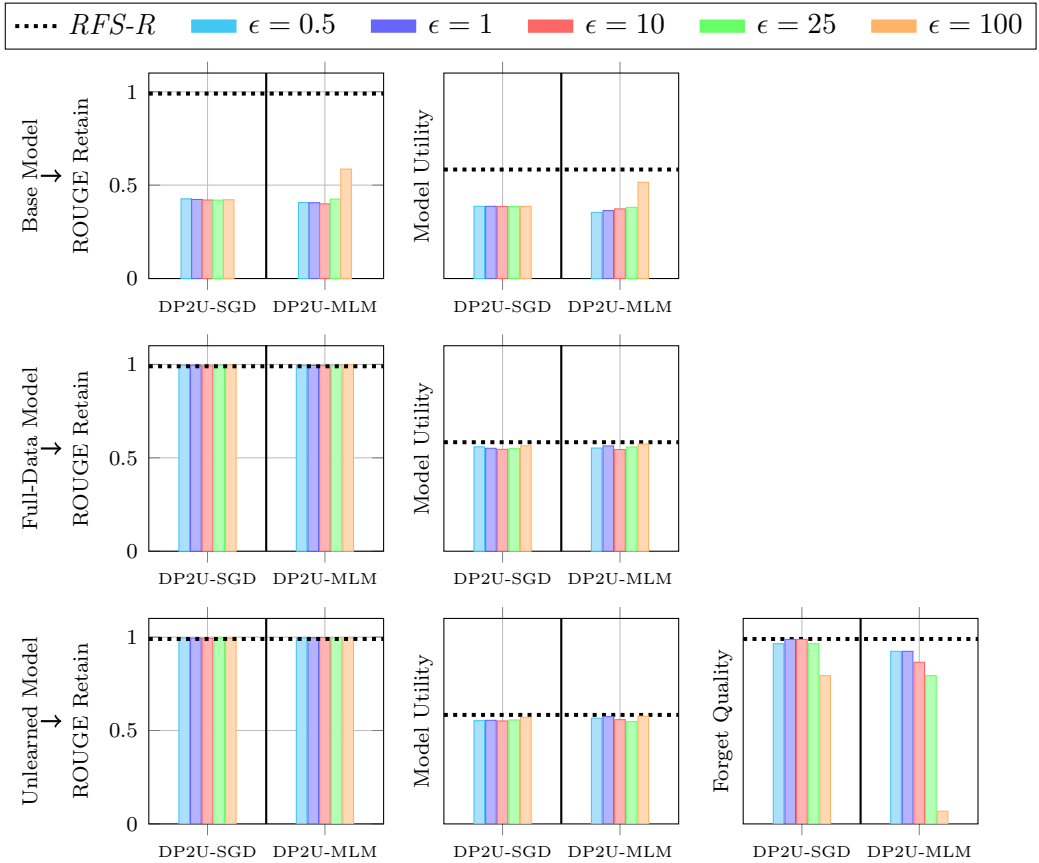
\begin{figure}[ht!]
    \centering
    \begin{tikzpicture}
        \begin{axis}[
            hide axis,
            width=1\textwidth, 
            height=2.5cm,
            xmin=0.5, xmax=1, 
            ymin=0.5, ymax=1, 
            legend columns=7,
            legend style={at={(0.5,1)}, anchor=north, /tikz/every even column/.append style={column sep=0.3cm}}
        ]
        \addlegendimage{dotted,thick,color=black, line width=1.5pt}
            \addlegendentry{\small{\textit{RFS-R}}}

            \addplot[color=cyan!60, fill=cyan!30, line width=5pt] coordinates {(0,0)};
            \addlegendentry{\small{\(\epsilon=0.5\)}}
            \addplot[color=blue!60, fill=blue!30, line width=5pt] coordinates {(0,0)};
            \addlegendentry{\small{\(\epsilon=1\)}}
            \addplot[color=red!60, fill=red!30, line width=5pt] coordinates {(0,0)};
            \addlegendentry{\small{\(\epsilon=10\)}}
            \addplot[color=green!60, fill=green!30, line width=5pt] coordinates {(0,0)};
            \addlegendentry{\small{\(\epsilon=25\)}}
            \addplot[color=orange!60, fill=orange!30, line width=5pt] coordinates {(0,0)};
            \addlegendentry{\small{\(\epsilon=100\)}}
        \end{axis}
    \end{tikzpicture}
    
    \vspace{-0.3cm}
    \begin{minipage}[t]{0.36\textwidth}
        \centering
        \begin{tikzpicture}
            \begin{axis}[
                width=0.95\textwidth, height=4.3cm, 
                xlabel={},
                ylabel={\parbox[c]{3cm}{\centering \scriptsize{Base Model\\\tikz{\draw[->,thick] (0,0) -- (0,-0.1);} \\ ROUGE Retain}}},
                symbolic x coords={DP-SGD, DP-MLM},
                xticklabels={\tiny{DP2U-SGD}, \tiny{DP2U-MLM}},
                xtick=data,
                ymin=0, ymax=1.1,
                ybar=0.0,
                bar width=0.14cm,
                enlarge x limits=0.5,
                grid=both,
                yticklabel style={font=\scriptsize}
            ]
\draw[thick, black, line width=0.8] ({rel axis cs:0.5,0}) -- ({rel axis cs:0.5,1.1});
            
\draw[dotted, thick, black, line width=1.5] ({rel axis cs:0,0.9}) -- ({rel axis cs:1,0.9});

\addplot[color=cyan!60, fill=cyan!30] coordinates {(DP-SGD, 0.4267) (DP-MLM, 0.4070)};
\addplot[color=blue!60, fill=blue!30] coordinates {(DP-SGD, 0.423334714) (DP-MLM, 0.4061)};
\addplot[color=red!60, fill=red!30] coordinates {(DP-SGD, 0.42047781) (DP-MLM, 0.4000)};
\addplot[color=green!60, fill=green!30] coordinates {(DP-SGD, 0.419573113) (DP-MLM, 0.425225284)};
\addplot[color=orange!60, fill=orange!30] coordinates {(DP-SGD, 0.421771416) (DP-MLM, 0.585839446)};
            \end{axis}
        \end{tikzpicture}
    \end{minipage}
    \hspace{-0.65cm}
\begin{minipage}[t]{0.36\textwidth}
        \centering
        \begin{tikzpicture}
            \begin{axis}[
                width=0.95\textwidth, height=4.3cm, 
                xlabel={},
                ylabel={\scriptsize{Model Utility}},
                symbolic x coords={DP-SGD, DP-MLM},
                xticklabels={\tiny{DP2U-SGD}, \tiny{DP2U-MLM}},
                xtick=data,
                ymin=0, ymax=1.0,
                ybar=0.4,
                bar width=0.14cm,
                enlarge x limits=0.5,
                grid=both,
                ytick=\empty
            ]
\draw[thick, black, line width=0.8] ({rel axis cs:0.5,0}) -- ({rel axis cs:0.5,1.1});

\draw[dotted, thick, black, line width=1.5] ({rel axis cs:0,0.5305}) -- ({rel axis cs:1,0.5305});

\addplot[color=cyan!60, fill=cyan!30]
  coordinates {(DP-SGD, 0.3514) (DP-MLM, 0.3219)};
\addplot[color=blue!60, fill=blue!30]
  coordinates {(DP-SGD, 0.3514) (DP-MLM, 0.3308)};
\addplot[color=red!60, fill=red!30]
  coordinates {(DP-SGD, 0.3510) (DP-MLM, 0.3389)};
\addplot[color=green!60, fill=green!30]
  coordinates {(DP-SGD, 0.3510) (DP-MLM, 0.3462)};
\addplot[color=orange!60, fill=orange!30]
  coordinates {(DP-SGD, 0.3513) (DP-MLM, 0.4685)};
            \end{axis}
        \end{tikzpicture}
    \end{minipage}  
    \hspace{-1.25cm}
        \begin{minipage}[t]{0.36\textwidth}
        \centering
        \begin{tikzpicture}[opacity=0]
            \begin{axis}[
                width=0.95\textwidth, height=4.3cm, 
                xlabel={},
                ylabel={\parbox[c]{3cm}{\centering \scriptsize{Base Model\\\tikz{\draw[->,thick] (0,0) -- (0,-0.1);} \\ ROUGE Retain}}},
                symbolic x coords={DP-SGD, DP-MLM},
                xticklabels={\tiny{DP2U-SGD}, \tiny{DP2U-MLM}},
                xtick=data,
                ymin=0, ymax=1.1,
                ybar=0.0,
                bar width=0.14cm,
                enlarge x limits=0.5,
                grid=both,
                yticklabel style={font=\scriptsize}
            ]
\addplot[color=gray!60, fill=gray!60] coordinates {(DP-SGD, 0) (DP-MLM, 0)};
          \end{axis}
        \end{tikzpicture}
    \end{minipage}
\vspace{0cm}
\begin{minipage}[t]{0.36\textwidth}
        \centering
        \begin{tikzpicture}
            \begin{axis}[
                width=0.95\textwidth, height=4.3cm, 
                xlabel={},
                ylabel={\parbox[c]{3cm}{\centering \scriptsize{Full-Data Model\\\tikz{\draw[->,thick] (0,0) -- (0,-0.1);} \\ ROUGE Retain}}},
                symbolic x coords={DP-SGD, DP-MLM},
                xticklabels={\tiny{DP2U-SGD}, \tiny{DP2U-MLM}},
                xtick=data,
                ymin=0, ymax=1.1,
                ybar=0.4,
                bar width=0.14cm,
                enlarge x limits=0.5,
                grid=both, 
                yticklabel style={font=\scriptsize}
            ]
\draw[thick, black, line width=0.8] ({rel axis cs:0.5,0}) -- ({rel axis cs:0.5,1.1});
            
\addplot[color=cyan!60, fill=cyan!30] coordinates {(DP-SGD, 0.9947) (DP-MLM, 0.9944)};
\addplot[color=blue!60, fill=blue!30] coordinates {(DP-SGD, 0.9957) (DP-MLM, 0.9946)};
\addplot[color=red!60, fill=red!30] coordinates {(DP-SGD, 0.9928) (DP-MLM, 0.9935)};
\addplot[color=green!60, fill=green!30] coordinates {(DP-SGD, 0.9945) (DP-MLM, 0.9935)};
\addplot[color=orange!60, fill=orange!30] coordinates {(DP-SGD, 0.9965) (DP-MLM, 0.9981)};

\draw[dotted, thick, black, line width=1.5] ({rel axis cs:0,0.9}) -- ({rel axis cs:1,0.9});
            \end{axis}
        \end{tikzpicture}
    \end{minipage}
    \hspace{-0.65cm}
    \begin{minipage}[t]{0.36\textwidth}
        \centering
        \begin{tikzpicture}
            \begin{axis}[
                width=0.95\textwidth, height=4.3cm, 
                xlabel={},
                ylabel={\scriptsize{Model Utility}},
                symbolic x coords={DP-SGD, DP-MLM},
                xticklabels={\tiny{DP2U-SGD}, \tiny{DP2U-MLM}},
                xtick=data,
                ymin=0, ymax=1.0,
                ybar=0.4,
                bar width=0.14cm,
                enlarge x limits=0.5,
                grid=both,
                ytick=\empty
            ]
\draw[thick, black, line width=0.8] ({rel axis cs:0.5,0}) -- ({rel axis cs:0.5,1.1});
            
\draw[dotted, thick, black, line width=1.5] ({rel axis cs:0,0.5305}) -- ({rel axis cs:1,0.5305});

\addplot[color=cyan!60, fill=cyan!30] coordinates {(DP-SGD, 0.5080) (DP-MLM, 0.5022)};
\addplot[color=blue!60, fill=blue!30] coordinates {(DP-SGD, 0.5010) (DP-MLM, 0.5124)};
\addplot[color=red!60, fill=red!30] coordinates {(DP-SGD, 0.4957) (DP-MLM, 0.4948)};
\addplot[color=green!60, fill=green!30] coordinates {(DP-SGD, 0.4991) (DP-MLM, 0.5071)};
\addplot[color=orange!60, fill=orange!30] coordinates {(DP-SGD, 0.5128) (DP-MLM, 0.5216)};
            \end{axis}
        \end{tikzpicture}
    \end{minipage}
        \hspace{-1.25cm}
        \begin{minipage}[t]{0.36\textwidth}
        \centering
        \begin{tikzpicture}[opacity=0]
            \begin{axis}[
                width=0.95\textwidth, height=4.3cm, 
                xlabel={},
                ylabel={\parbox[c]{3cm}{\centering \scriptsize{Base Model\\\tikz{\draw[->,thick] (0,0) -- (0,-0.1);} \\ ROUGE Retain}}},
                symbolic x coords={DP-SGD, DP-MLM},
                xticklabels={\tiny{DP2U-SGD}, \tiny{DP2U-MLM}},
                xtick=data,
                ymin=0, ymax=1.1,
                ybar=0.0,
                bar width=0.14cm,
                enlarge x limits=0.5,
                grid=both,
                yticklabel style={font=\scriptsize}
            ]
\addplot[color=gray!60, fill=gray!60] coordinates {(DP-SGD, 0) (DP-MLM, 0)};
          \end{axis}
        \end{tikzpicture}
    \end{minipage}
\vspace{0cm}
\begin{minipage}[t]{0.36\textwidth}
        \centering
        \begin{tikzpicture}
            \begin{axis}[
                width=0.95\textwidth, height=4.3cm, 
                xlabel={},
                ylabel={\parbox[c]{3cm}{\centering \scriptsize{Unlearned Model\\\tikz{\draw[->,thick] (0,0) -- (0,-0.1);} \\ ROUGE Retain}}},
                symbolic x coords={DP-SGD, DP-MLM},
                xticklabels={\tiny{DP2U-SGD}, \tiny{DP2U-MLM}},
                xtick=data,
                ymin=0, ymax=1.1,
                ybar=0.4,
                bar width=0.14cm,
                enlarge x limits=0.5,
                grid=both, 
                yticklabel style={font=\scriptsize}
            ]
\draw[thick, black, line width=0.8] ({rel axis cs:0.5,0}) -- ({rel axis cs:0.5,1.1});
            
\addplot[color=cyan!60, fill=cyan!30] coordinates {(DP-SGD, 0.9956) (DP-MLM, 0.9958)};
\addplot[color=blue!60, fill=blue!30] coordinates {(DP-SGD, 0.9943) (DP-MLM, 0.9964)};
\addplot[color=red!60, fill=red!30] coordinates {(DP-SGD, 0.9935) (DP-MLM, 0.9976)};
\addplot[color=green!60, fill=green!30] coordinates {(DP-SGD, 0.9944) (DP-MLM, 0.9963)};
\addplot[color=orange!60, fill=orange!30] coordinates {(DP-SGD, 0.9930) (DP-MLM, 0.9962)};
\draw[dotted, thick, black, line width=1.5] ({rel axis cs:0,0.9}) -- ({rel axis cs:1,0.9});
            \end{axis}
        \end{tikzpicture}
    \end{minipage}
    \hspace{-0.65cm}
    \begin{minipage}[t]{0.36\textwidth}
        \centering
        \begin{tikzpicture}
            \begin{axis}[
                width=0.95\textwidth, height=4.3cm, 
                xlabel={},
                ylabel={\scriptsize{Model Utility}},
                symbolic x coords={DP-SGD, DP-MLM},
                xticklabels={\tiny{DP2U-SGD}, \tiny{DP2U-MLM}},
                xtick=data,
                ymin=0, ymax=1.0,
                ybar=0.4,
                bar width=0.14cm,
                enlarge x limits=0.5,
                grid=both,
                ytick=\empty
            ]
\draw[thick, black, line width=0.8] ({rel axis cs:0.5,0}) -- ({rel axis cs:0.5,1.1});
            
\draw[dotted, thick, black, line width=1.5] ({rel axis cs:0,0.5305}) -- ({rel axis cs:1,0.5305});

\addplot[color=cyan!60, fill=cyan!30] coordinates {(DP-SGD, 0.5023) (DP-MLM, 0.5146)};
\addplot[color=blue!60, fill=blue!30] coordinates {(DP-SGD, 0.5042) (DP-MLM, 0.5223)};
\addplot[color=red!60, fill=red!30] coordinates {(DP-SGD, 0.5011) (DP-MLM, 0.5074)};
\addplot[color=green!60, fill=green!30] coordinates {(DP-SGD, 0.5059) (DP-MLM, 0.4980)};
\addplot[color=orange!60, fill=orange!30] coordinates {(DP-SGD, 0.5188) (DP-MLM, 0.5241)};
            \end{axis}
        \end{tikzpicture}
    \end{minipage}
    \hspace{-1.25cm}
    \begin{minipage}[t]{0.36\textwidth}
        \centering
        \begin{tikzpicture}
            \begin{axis}[
                width=0.95\textwidth, height=4.3cm,
                xlabel={},
                ylabel={\scriptsize{Forget Quality}},
                symbolic x coords={DP-SGD, DP-MLM},
                xticklabels={\tiny{DP2U-SGD}, \tiny{DP2U-MLM}},
                xtick=data,
                ymin=0, ymax=1.1,
                ybar=0.4,
                bar width=0.14cm,
                enlarge x limits=0.5,
                grid=both,
                ytick=\empty
            ]
\draw[thick, black, line width=0.8] ({rel axis cs:0.5,0}) -- ({rel axis cs:0.5,1.1});
            
\draw[dotted, thick, black, line width=1.5] ({rel axis cs:0,0.9}) -- ({rel axis cs:1,0.9});

\addplot[color=cyan!60, fill=cyan!30] coordinates {(DP-SGD, 0.9647) (DP-MLM, 0.9238)};
\addplot[color=blue!60, fill=blue!30] coordinates {(DP-SGD, 0.9878) (DP-MLM, 0.9238)};
\addplot[color=red!60, fill=red!30] coordinates {(DP-SGD, 0.9878) (DP-MLM, 0.8655)};
\addplot[color=green!60, fill=green!30] coordinates {(DP-SGD, 0.9647) (DP-MLM, 0.7934)};
\addplot[color=orange!60, fill=orange!30] coordinates {(DP-SGD, 0.7934) (DP-MLM, 0.0680)};
            \end{axis}
        \end{tikzpicture}
    \end{minipage}

    \vspace{-0.2cm}
    \caption{Evaluation results for Phi and 5\% forget ratio. $E=10$ for \textit{RFS-R} (Retrain from scratch on the set to be retained) and $E'=5$ for DP2U-SGD and DP2U-MLM for all $\epsilon$ values.}
\label{figure3}
\end{figure}

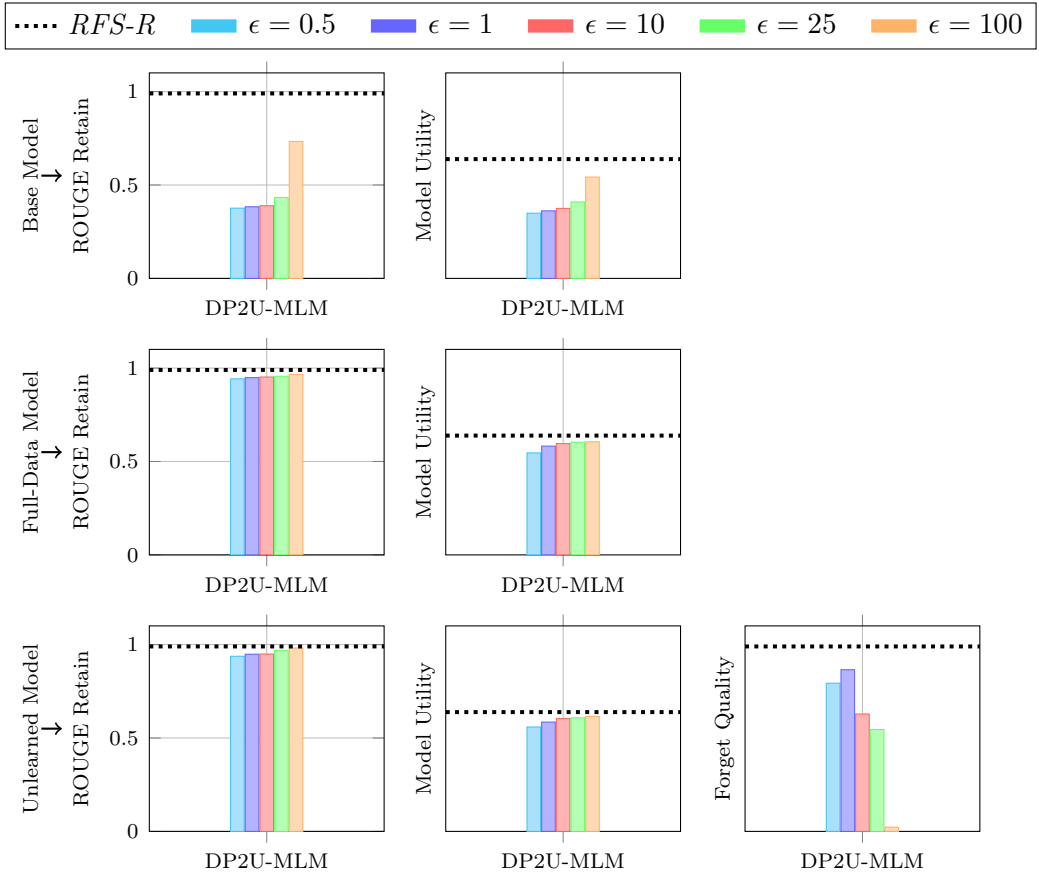
\begin{figure}[ht!]
    \centering
    \begin{tikzpicture}
        \begin{axis}[
            hide axis,
            width=1\textwidth, 
            height=2.5cm,
            xmin=0.5, xmax=1, 
            ymin=0.5, ymax=1, 
            legend columns=7,
            legend style={at={(0.5,1)}, anchor=north, /tikz/every even column/.append style={column sep=0.3cm}}
        ]
\addlegendimage{dotted,thick,color=black, line width=1.5pt}
            \addlegendentry{\small{\textit{RFS-R}}}
            \addplot[color=cyan!60, fill=cyan!30, line width=5pt] coordinates {(0,0)};
            \addlegendentry{\small{\(\epsilon=0.5\)}}
            \addplot[color=blue!60, fill=blue!30, line width=5pt] coordinates {(0,0)};
            \addlegendentry{\small{\(\epsilon=1\)}}
            \addplot[color=red!60, fill=red!30, line width=5pt] coordinates {(0,0)};
            \addlegendentry{\small{\(\epsilon=10\)}}
            \addplot[color=green!60, fill=green!30, line width=5pt] coordinates {(0,0)};
            \addlegendentry{\small{\(\epsilon=25\)}}
            \addplot[color=orange!60, fill=orange!30, line width=5pt] coordinates {(0,0)};
            \addlegendentry{\small{\(\epsilon=100\)}}
        \end{axis}
    \end{tikzpicture}
    
    \vspace{-0.3cm}
\begin{minipage}[t]{0.36\textwidth}
    \centering
    \begin{tikzpicture}
        \begin{axis}[
            width=0.95\textwidth, height=4.3cm, 
            xlabel={},
            ylabel={\parbox[c]{3cm}{\centering \scriptsize{Base Model\\\tikz{\draw[->,thick] (0,0) -- (0,-0.1);} \\ ROUGE Retain}}},
            symbolic x coords={Left, DP-MLM, Right},
            xticklabels={\scriptsize{DP2U-MLM}},
            xtick=data,
            xmin=Left, xmax=Right,
            ymin=0, ymax=1.1,
            ybar=0.4,
            bar width=0.18cm,
            grid=both,
            yticklabel style={font=\scriptsize}
        ]
\draw[dotted, thick, black, line width=1.5] ({rel axis cs:0,0.9}) -- ({rel axis cs:1,0.9});

\addplot[color=cyan!60, fill=cyan!30] coordinates {(DP-MLM, 0.3760)};
\addplot[color=blue!60, fill=blue!30] coordinates {(DP-MLM, 0.3834)};
\addplot[color=red!60, fill=red!30] coordinates {(DP-MLM, 0.3892)};
\addplot[color=green!60, fill=green!30] coordinates {(DP-MLM, 0.4331)};
\addplot[color=orange!60, fill=orange!30] coordinates {(DP-MLM, 0.7330)};
        \end{axis}
    \end{tikzpicture}
\end{minipage}
    \hspace{-0.65cm}
\begin{minipage}[t]{0.36\textwidth}
        \centering
        \begin{tikzpicture}
            \begin{axis}[
                width=0.95\textwidth, height=4.3cm, 
                xlabel={},
                ylabel={\scriptsize{Model Utility}},
            symbolic x coords={Left, DP-MLM, Right},
            xticklabels={\scriptsize{DP2U-MLM}},
            xtick=data,
            xmin=Left, xmax=Right,
            ymin=0, ymax=1.0,
            ybar=0.4,
            bar width=0.18cm,
            grid=both,
                ytick=\empty
            ]
\draw[dotted, thick, black, line width=1.5] ({rel axis cs:0,0.5805}) -- ({rel axis cs:1,0.5805});

\addplot[color=cyan!60, fill=cyan!30] coordinates {(DP-MLM, 0.3173)};
\addplot[color=blue!60, fill=blue!30] coordinates {(DP-MLM, 0.3286)};
\addplot[color=red!60, fill=red!30] coordinates {(DP-MLM, 0.3406)};
\addplot[color=green!60, fill=green!30] coordinates {(DP-MLM, 0.3718)};
\addplot[color=orange!60, fill=orange!30] coordinates {(DP-MLM, 0.4931)};

            \end{axis}
        \end{tikzpicture}
    \end{minipage}   
\hspace{-1.25cm}
\begin{minipage}[t]{0.36\textwidth}
        \centering
        \begin{tikzpicture}[opacity=0]
            \begin{axis}[
                width=0.95\textwidth, height=4.3cm,
                xlabel={},
                ylabel={\scriptsize{Forget Quality}},
            symbolic x coords={Left, DP-MLM, Right},
            xticklabels={\scriptsize{DP2U-MLM}},
            xtick=data,
            xmin=Left, xmax=Right,
            ymin=0, ymax=1.1,
            ybar=0.4,
            bar width=0.18cm,
            grid=both,
                ytick=\empty
            ]
\addplot[color=orange!60, fill=orange!30] coordinates {(DP-MLM, 0)};
            \end{axis}
        \end{tikzpicture}
    \end{minipage}
    
\vspace{0cm}
\begin{minipage}[t]{0.36\textwidth}
        \centering
        \begin{tikzpicture}
            \begin{axis}[
                width=0.95\textwidth, height=4.3cm, 
                xlabel={},
                ylabel={\parbox[c]{3cm}{\centering \scriptsize{Full-Data Model\\\tikz{\draw[->,thick] (0,0) -- (0,-0.1);} \\ ROUGE Retain}}},
            symbolic x coords={Left, DP-MLM, Right},
            xticklabels={\scriptsize{DP2U-MLM}},
            xtick=data,
            xmin=Left, xmax=Right,
            ymin=0, ymax=1.1,
            ybar=0.4,
            bar width=0.18cm,
            grid=both,
                yticklabel style={font=\scriptsize}
            ]
\addplot[color=cyan!60, fill=cyan!30] coordinates {(DP-MLM, 0.9423)};
\addplot[color=blue!60, fill=blue!30] coordinates {(DP-MLM, 0.9489)};
\addplot[color=red!60, fill=red!30] coordinates {(DP-MLM, 0.9522)};
\addplot[color=green!60, fill=green!30] coordinates {(DP-MLM, 0.9564)};
\addplot[color=orange!60, fill=orange!30] coordinates {(DP-MLM, 0.9665)};

\draw[dotted, thick, black, line width=1.5] ({rel axis cs:0,0.9}) -- ({rel axis cs:1,0.9});
            \end{axis}
        \end{tikzpicture}
    \end{minipage}
    \hspace{-0.65cm}
    \begin{minipage}[t]{0.36\textwidth}
        \centering
        \begin{tikzpicture}
            \begin{axis}[
                width=0.95\textwidth, height=4.3cm, 
                xlabel={},
                ylabel={\scriptsize{Model Utility}},
            symbolic x coords={Left, DP-MLM, Right},
            xticklabels={\scriptsize{DP2U-MLM}},
            xtick=data,
            xmin=Left, xmax=Right,
            ymin=0, ymax=1.0,
            ybar=0.4,
            bar width=0.18cm,
            grid=both,
                ytick=\empty
            ]
\draw[dotted, thick, black, line width=1.5] ({rel axis cs:0,0.5805}) -- ({rel axis cs:1,0.5805});

\addplot[color=cyan!60, fill=cyan!30] coordinates {(DP-MLM, 0.4963)};
\addplot[color=blue!60, fill=blue!30] coordinates {(DP-MLM, 0.5294)};
\addplot[color=red!60, fill=red!30] coordinates {(DP-MLM, 0.5416)};
\addplot[color=green!60, fill=green!30] coordinates {(DP-MLM, 0.5485)};
\addplot[color=orange!60, fill=orange!30] coordinates {(DP-MLM, 0.5504)};

            \end{axis}
        \end{tikzpicture}
    \end{minipage}
    \hspace{-1.25cm}
    \begin{minipage}[t]{0.36\textwidth}
        \centering
        \begin{tikzpicture}[opacity=0]
            \begin{axis}[
                width=0.95\textwidth, height=4.3cm,
                xlabel={},
                ylabel={\scriptsize{Forget Quality}},
            symbolic x coords={Left, DP-MLM, Right},
            xticklabels={\scriptsize{DP2U-MLM}},
            xtick=data,
            xmin=Left, xmax=Right,
            ymin=0, ymax=1.1,
            ybar=0.4,
            bar width=0.18cm,
            grid=both,
                ytick=\empty
            ]
\addplot[color=orange!60, fill=orange!30] coordinates {(DP-MLM,0)};

            \end{axis}
        \end{tikzpicture}
    \end{minipage}

\vspace{0cm}
\begin{minipage}[t]{0.36\textwidth}
        \centering
        \begin{tikzpicture}
            \begin{axis}[
                width=0.95\textwidth, height=4.3cm, 
                xlabel={},
                ylabel={\parbox[c]{3cm}{\centering \scriptsize{Unlearned Model\\\tikz{\draw[->,thick] (0,0) -- (0,-0.1);} \\ ROUGE Retain}}},
            symbolic x coords={Left, DP-MLM, Right},
            xticklabels={\scriptsize{DP2U-MLM}},
            xtick=data,
            xmin=Left, xmax=Right,
            ymin=0, ymax=1.1,
            ybar=0.4,
            bar width=0.18cm,
            grid=both,
                yticklabel style={font=\scriptsize}
            ]

\addplot[color=cyan!60, fill=cyan!30] coordinates {(DP-MLM, 0.9374)};
\addplot[color=blue!60, fill=blue!30] coordinates {(DP-MLM, 0.9484)};
\addplot[color=red!60, fill=red!30] coordinates {(DP-MLM, 0.9494)};
\addplot[color=green!60, fill=green!30] coordinates {(DP-MLM, 0.9692)};
\addplot[color=orange!60, fill=orange!30] coordinates {(DP-MLM, 0.9813)};

\draw[dotted, thick, black, line width=1.5] ({rel axis cs:0,0.9}) -- ({rel axis cs:1,0.9});
            \end{axis}
        \end{tikzpicture}
    \end{minipage}
    \hspace{-0.65cm}
    \begin{minipage}[t]{0.36\textwidth}
        \centering
        \begin{tikzpicture}
            \begin{axis}[
                width=0.95\textwidth, height=4.3cm, 
                xlabel={},
                ylabel={\scriptsize{Model Utility}},
            symbolic x coords={Left, DP-MLM, Right},
            xticklabels={\scriptsize{DP2U-MLM}},
            xtick=data,
            xmin=Left, xmax=Right,
            ymin=0, ymax=1.0,
            ybar=0.4,
            bar width=0.18cm,
            grid=both,
                ytick=\empty
            ]
\draw[dotted, thick, black, line width=1.5] ({rel axis cs:0,0.5805}) -- ({rel axis cs:1,0.5805});

\addplot[color=cyan!60, fill=cyan!30] coordinates {(DP-MLM, 0.5080)};
\addplot[color=blue!60, fill=blue!30] coordinates {(DP-MLM, 0.5320)};
\addplot[color=red!60, fill=red!30] coordinates {(DP-MLM, 0.5488)};
\addplot[color=green!60, fill=green!30] coordinates {(DP-MLM, 0.5526)};
\addplot[color=orange!60, fill=orange!30] coordinates {(DP-MLM, 0.5594)};

            \end{axis}
        \end{tikzpicture}
    \end{minipage}
    \hspace{-1.25cm}
    \begin{minipage}[t]{0.36\textwidth}
        \centering
        \begin{tikzpicture}
            \begin{axis}[
                width=0.95\textwidth, height=4.3cm,
                xlabel={},
                ylabel={\scriptsize{Forget Quality}},
            symbolic x coords={Left, DP-MLM, Right},
            xticklabels={\scriptsize{DP2U-MLM}},
            xtick=data,
            xmin=Left, xmax=Right,
            ymin=0, ymax=1.1,
            ybar=0.4,
            bar width=0.18cm,
            grid=both,
                ytick=\empty
            ]
\draw[dotted, thick, black, line width=1.5] ({rel axis cs:0,0.9}) -- ({rel axis cs:1,0.9});

\addplot[color=cyan!60, fill=cyan!30] coordinates {(DP-MLM, 0.7934)};
\addplot[color=blue!60, fill=blue!30] coordinates {(DP-MLM, 0.8655)};
\addplot[color=red!60, fill=red!30] coordinates {(DP-MLM, 0.6284)};
\addplot[color=green!60, fill=green!30] coordinates {(DP-MLM, 0.5453)};
\addplot[color=orange!60, fill=orange!30] coordinates {(DP-MLM, 0.023)};

            \end{axis}
        \end{tikzpicture}
    \end{minipage}
    
    \vspace{-0.2cm}
    \caption{Evaluation results for Llama2 and 5\% forget ratio. $E=10$ for \textit{RFS-R} and $E'=5$ for DP2U-MLM for all epsilon values.}
    \label{figure4}
\end{figure}

We analyze how varying the values of \(\epsilon\) (\emph{i.e.}, $0.5, 1, 10, 25, 100$) affected ROUGE, the utility of the model, and the quality of forgetting. Figures~\ref{figure3} and \ref{figure4} show the results for the Phi and Llama2 models, respectively, across three model states: the base model (trained with DP), the full data model (fine-tuned with full original data) and the unlearned model (fine-tuned only on the data to be retained).



We observe the following.

\textbf{DP2U-SGD:} On Phi (the computational constraints of DP-SGD prevent training Llama2 within our setup), DP2U-SGD maintains consistency in all metrics regardless of the different values of $\epsilon$. Generally, larger $\epsilon$ values enhance model utility by compromising disclosure restrictions, whereas smaller $\epsilon$ values provide better disclosure protection, albeit with a reduction in utility. However, consistent results are mainly due to the DP-SGD mechanism of controlled noise addition and gradient clipping, which guarantees that updates are limited and are quite insensitive to values $\epsilon$. As a result, while the ROUGE retain and model utility metrics in the unlearned model improve and closely match those of RFS-R, the values remain nearly identical across all $\epsilon$ levels for all model states. This aligns with the theory that DP-SGD protects disclosure information by limiting individual gradient contributions rather than completely removing learned information.

\textbf{DP2U-MLM: } Unlike DP-SGD, DP-MLM relies on replacing disclosive terms (mainly noun phrases) with probabilistically chosen tokens, which makes it fairly affected by \(\epsilon\). This effect is particularly pronounced in larger models such as Llama2. As $\epsilon$ increases, the utility of the model increases due to fewer token substitutions, allowing the model to retain the semantic structure better. In contrast, at lower $\epsilon$ values, excessive token replacements degrade the learned representations, and fine-tuning struggles to fully restore model utility. Interestingly, this pattern is not as pronounced in the smaller Phi model, which shows fairly consistent performance at various values of $\epsilon$. Given that most LLMs are more similar to Llama2 in size and architecture than Phi, these findings imply that the sensitivity observed to $\epsilon$ is likely to be relative to a wider range of models. Despite these effects, in the unlearned model, DP-MLM demonstrates enhanced performance in all metrics and brings them close to the benchmark RFS-R. Importantly, while the utility of the model shows minimal change at various values of \(\epsilon\), the forget quality shows considerable variation. This suggests that the choice of $\epsilon$ critically influences the balance between disclosure protection and model performance.

\textbf{Choosing $\epsilon$:} According to the recommendations of the inventors of DP (mentioned in Section \ref{sec3}), $\epsilon\leq1$ achieves strong disclosure protection, which is critical for \emph{guaranteed} unlearning in our approaches. However, setting $\epsilon$ should balance disclosure protection and model performance. These are our findings based on the results reported above:

\begin{itemize}
    \item \textbf{For $\epsilon<1$:} The base model reduces the utility of the model (more apparently in the larger model, Llama2) but provides greater disclosure protection after unlearning. This is closer to RFS-R, as it provides guaranteed protection against disclosure but requires more extensive fine-tuning to recover the utility.  
    \item \textbf{For $\epsilon>10$:} The base model retains higher model utility (more apparently in the larger model, Llama2), but may compromise disclosure protection after unlearning. This is closer to approximate unlearning, as less effort is required to restore the utility of the model by sacrificing unlearning guarantees.
\end{itemize}
Therefore, based on our findings, $\epsilon=1$ offers a good trade-off, as it provides theoretically and practically meaningful protection against disclosure (\emph{i.e.},  guaranteed unlearning) with manageable fine-tuning requirements. This result aligns with the recommendation of the DP inventors.

\subsection{Comparative analysis} 

Next, we compare our approach with exact unlearning (RFS-R) and approximate unlearning techniques (GA, GD, KL, and PO). As a reference, in Table~\ref{table2} we report the utility of the models before unlearning.
The evaluation results for the Retain and Forget sets are detailed in Table~\ref{table3} and depicted in Figures~\ref{figure6} and \ref{figure7}. Additional results for the four baseline data sets are included in~\ref{appendix:1}. 

\begin{table}[ht!]
\centering
\scriptsize 
\caption{Utility of the models before unlearning. FT-RF stands for fine-tuning on both retain and forget datasets.}
\begin{tabular}{|l|c|c|}
\hline
\textbf{Models} & \textbf{Pre-trained} & \textbf{Non-DP FT-RF} \\
\hline
Phi    & 0.3354 & 0.5411 \\
Llama2 & 0.2516 & 0.5793 \\
\hline
\end{tabular}
\label{table2}
\end{table}

For a fair comparison, DP2U-MLM and DP2U-SGD were evaluated using the chosen privacy budget $\epsilon=1$. Additional results for $\epsilon=0.5, 10, 25, 100$ are presented in \ref{appendix:4} and \ref{appendix:5}.

\subsubsection{Overall forget quality and model utility}
\label{subsubsec6.2.1}

Table~\ref{table3} presents the overall forget quality (FQ) and model utility (MU) across different unlearning methods for the Phi and Llama2 models, evaluated at three forget ratios (1\%, 5\%, and 10\%). These results highlight the trade-offs between retention and unlearning effectiveness in various approaches.

\begin{table}[ht!]
    \centering
    \scriptsize
    \caption{Forget quality and model utility of different unlearning methods for Phi and Llama2 at different forget ratios. For each metric, the best result is \textbf{bolded}, and the second-best is \underline{underlined}. \textit{RFS-R} serves as the benchmark but is excluded from ranking due to its high computational cost. Our methods are highlighted by gray cell color.}
    \renewcommand{\arraystretch}{1.3}
    \begin{tabular}{l|l|c|cc|cc|cc}
        \toprule
        \multirow{4}{*}{Model} & \multirow{4}{*}{Method} & \multirow{4}{*}{Epochs} & \multicolumn{6}{c}{Forget Ratios} \\
        \cmidrule(lr){4-9}
        & & & \multicolumn{2}{|c|}{1\%} & \multicolumn{2}{|c|}{5\%} & \multicolumn{2}{|c}{10\%} \\
        \cmidrule(lr){4-5} \cmidrule(lr){6-7} \cmidrule(lr){8-9}
        & & $\downarrow$& MU $\uparrow$ & FQ $\uparrow$ & MU $\uparrow$ & FQ $\uparrow$ & MU $\uparrow$ & FQ $\uparrow$ \\
        \midrule
        \multirow{7}{*}{Phi} &RFS-R & 10 & 0.5448 & 1.0000 & 0.5380 & 1.0000 & 0.5314 & 1.0000 \\
        & GA & 5 & 0.0230 & 0.0143 & 0.0000 & 0.0021 & 0.0000 & 8.84E-08 \\
        & GD & 5 & 0.4329 & \underline{0.1650} & 0.1982 & 1.87E-09 & 0.4005 & 5.56E-14 \\
        & KL & 5 & 0.0210 & 0.0143 & 0.0000 & 0.0021 & 0.0000 & 1.46E-14 \\
        & PO & 5 & \textbf{0.5223} & 0.0013 & {0.5114} & 2.56E-14 & \textbf{0.5313} & 7.90E-22 \\
        & \cellcolor{gray!30}DP2U-SGD & \cellcolor{gray!30}5 & \cellcolor{gray!30}\underline{0.5060} & \cellcolor{gray!30}\textbf{0.9900} & \cellcolor{gray!30}\underline{0.5122} & \cellcolor{gray!30}\textbf{0.9878} & \cellcolor{gray!30}0.5113 & \cellcolor{gray!30}\underline{0.9003} \\
        & \cellcolor{gray!30}DP2U-MLM & \cellcolor{gray!30}5 & \cellcolor{gray!30}0.5026 & \cellcolor{gray!30}\textbf{0.9900} & \cellcolor{gray!30}\textbf{0.5223} & \cellcolor{gray!30}\underline{0.9238} & \cellcolor{gray!30}\underline{0.5134} & \cellcolor{gray!30}\textbf{0.9014} \\
        \midrule
        \multirow{6}{*}{Llama2} & RFS-R & 6 & 0.5870 & 1.0000 & 0.5711 & 1.0000 & 0.5688 & 1.0000 \\
        & GA & 3 & 0.0000 & \underline{0.7659} & 0.0000 & 2.61E-07 & 0.0000 & 1.85E-15 \\
        & GD & 3 & 0.0000 & 0.2657 & 0.3490 & 1.39E-11 & 0.4053 & 2.86E-14 \\
        & KL & 3 & 0.0000 & 0.4046 & 0.0000 & \underline{4.61E-07} & 0.0000 & \underline{2.59E-12} \\
        & PO & 3 & \underline{0.4905} & 0.0013 & \underline{0.4950} & 1.83E-19 & \underline{0.5290} & 2.43E-19 \\
        & \cellcolor{gray!30}DP2U-MLM &\cellcolor{gray!30}3 &\cellcolor{gray!30}\textbf{0.5231} &\cellcolor{gray!30}\textbf{0.9999} &\cellcolor{gray!30}\textbf{0.5320} &\cellcolor{gray!30}\textbf{0.8655} & \cellcolor{gray!30}\textbf{0.5378} & \cellcolor{gray!30}\textbf{0.1761} \\
        \bottomrule
    \end{tabular}
    \label{table3}
\end{table}

With 1\% forgetting, GA, GD, KL, and PO achieve some varying degrees of forget quality and model utility. However, PO does not exceed the $0.05$ KS $p$ value threshold, indicating that it does not satisfy statistically significant forgetting. Furthermore, for GA, GD, and KL, the forget quality remains low for Phi, whereas Llama2 exhibits a trade-off between forget quality and model utility, suggesting that model size and architecture influence forgetting performance.
    
With larger forget ratios (5\% and 10\%), all approximate methods fail to achieve a meaningful forgetting (KS $p$-value $\geq0.05$). In particular, PO and GD exhibit some level of model utility but negligible forget quality, implying that they do not effectively remove unwanted information. In general, all approximate methods exhibit extremely poor forget quality at higher forget ratios, often approaching near zero. This highlights that these methods are ineffective for reliable long-term forgetting, especially when more data need to be forgotten.
    
Both DP2U-SGD and DP2U-MLM achieved results comparable to exact forgetting RFS-R, showcasing their robustness. Although DP2U-SGD maintains high model utility while ensuring strong forget quality, it requires significant computational resources, making it impractical for large models. In contrast, DP2U-MLM stands out as the best performing method for Phi and Llama2, balancing high utility (MU is 0.5222 for Phi and 0.5323 for Llama2, close to the MU before
forgetting reported in Table~\ref{table2}) and strong forgetting quality (FQ is 0.9238 for Phi and 0.8655 for Llama2), even at moderate forget ratios of 5\%. These results confirm that DP2U-MLM provides a compelling alternative to RFS-R at a fraction of the computational cost, making it ideal for scalable unlearning in resource-constrained environments.

\begin{figure}[ht]
    \centering
    \begin{minipage}{1\textwidth}
        \centering        \includegraphics[width=0.9\textwidth]{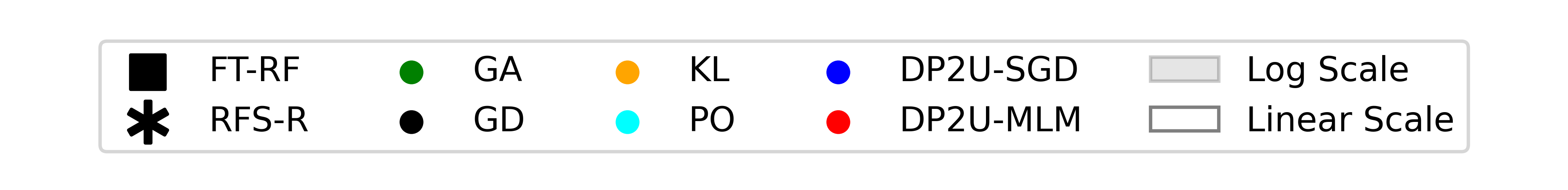}
    \end{minipage}
    
    \vspace{-0.2cm}
    \begin{minipage}{0.52\textwidth}
        \centering
\includegraphics[width=1\textwidth]{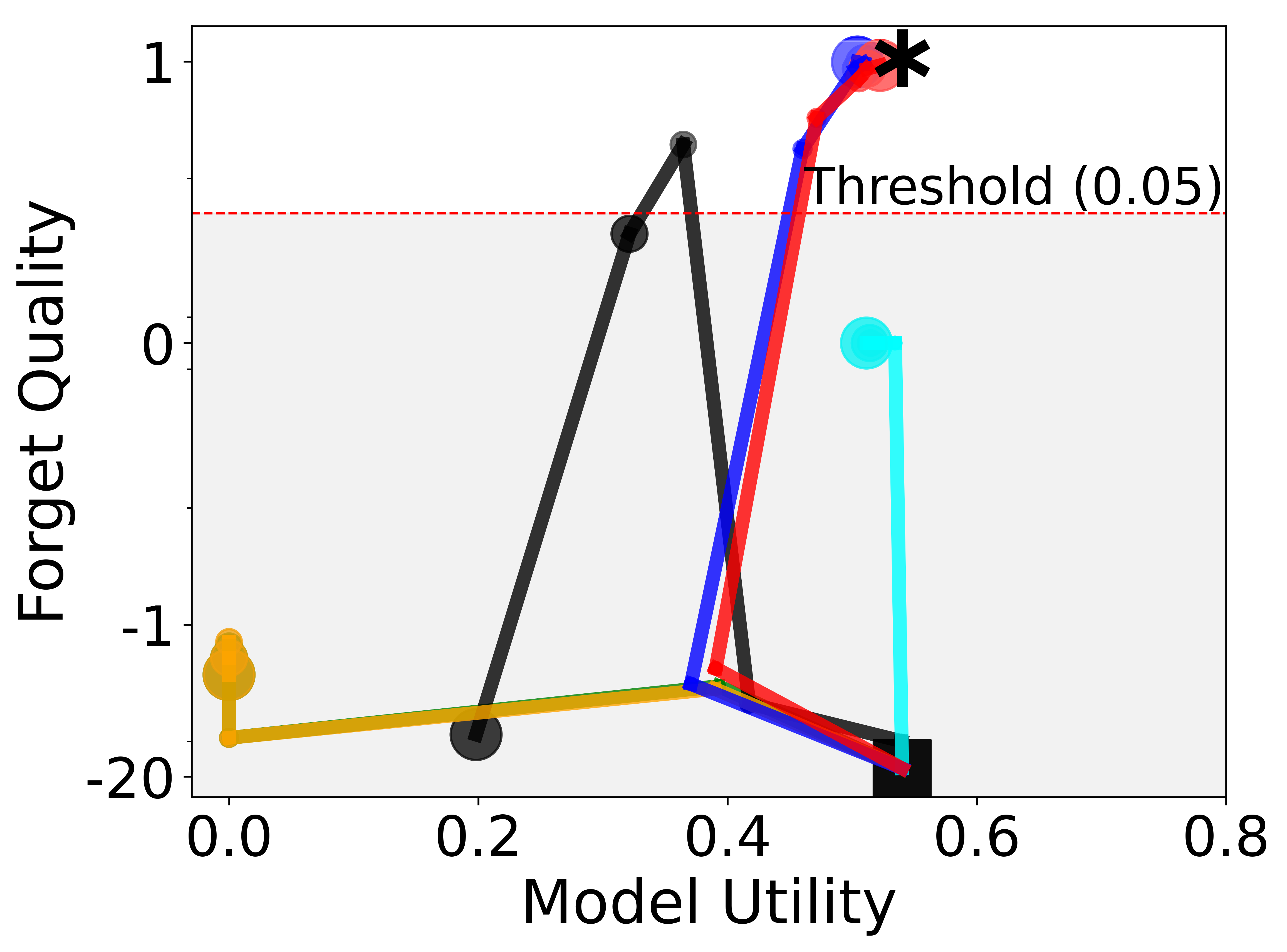}
    \end{minipage}
    \hspace{-0.3cm}
    \begin{minipage}{0.48\textwidth}
        \centering
\includegraphics[width=1\textwidth]{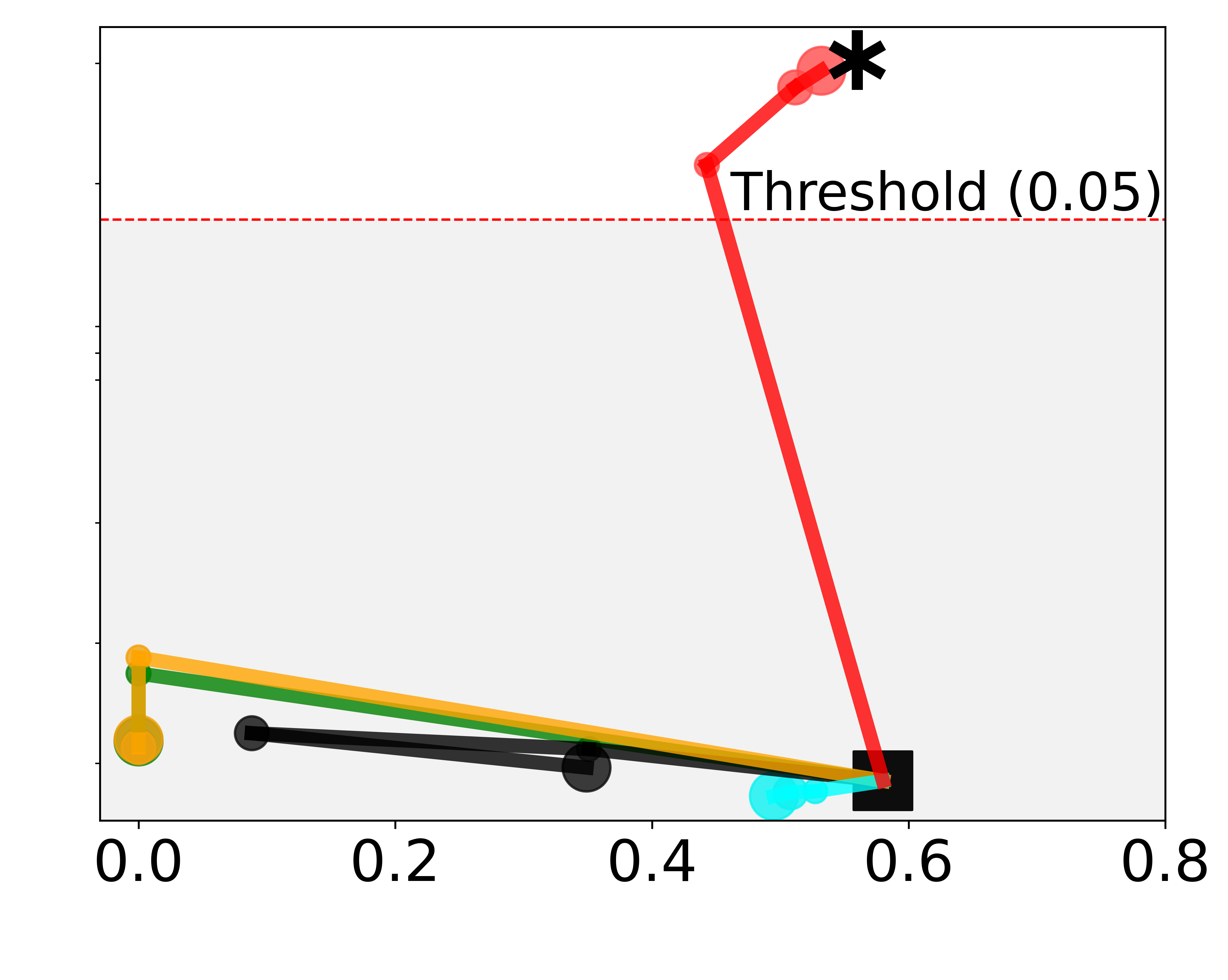}
    \end{minipage}
    
    \vspace{-0.1cm}
    \caption{Forget quality vs. model utility for the Phi (left) and Llama2 (right) across different unlearning methods and epochs. 
    The size of markers grows with the number of epochs.}
    \label{figure5}
\end{figure}

To gain a more comprehensive understanding of how the performance of the model changes over epochs, Figure~\ref{figure5} presents an in-depth comparison of the forget quality and the model utility for various unlearning techniques at 5\% forget ratio. The epoch-wise evolution for the 1\% and 10\% forget ratios is presented in~\ref{appendix:2} to offer a more thorough perspective on how the methods perform under different forget conditions.

Figure~\ref{figure5} illustrates the trade-off between forget quality and model utility across different unlearning techniques for Phi (left) and Llama2 (right) models. Each method is evaluated at multiple epochs, where the comparative size of the markers (circles, squares, and asterisks) represents the number of epochs. The larger the epoch number, the larger the marker size. The epochs are $E=10$ for FT-RF and RFS-R, while $E'=5$ for other methods.

It is evident from the results that, with the exception of GD at some epoch for the Phi model, all approximate techniques failed to achieve the minimum meaningful forgetting quality (\emph{i.e.}, a KS $p$-value above the threshold of $0.05$). Further, the behaviors of different unlearning methods across epochs are far from uniform. Fluctuations are particularly evident in approximate methods, where forget quality varies as epochs increase. These methods also display a clear trade-off with model utility: as the number of unlearning epochs increases, performance declines, causing utility to drop sharply (with trajectories shifting left), often approaching zero. In particular:

    \begin{itemize}
        \item[-] For GD, the fluctuations are related to the nature of gradient-based updates. In early epochs, when the loss is high, GD makes large updates, aggressively unlearning the forget set. However, as loss decreases, the gradients become smaller and can reverse the direction of updates, unintentionally reinforce forgotten information, and damage retention. This non-monotonic behavior is more pronounced in larger models like Llama2, where the complex loss landscape makes updates less predictable.
        
        \item [-] PO exhibits a more consistent trend in terms of model utility, albeit with a reduced forget quality. This happens because PO replaces responses on forgotten data with predetermined responses, which limits parameter adjustments and consequently preserves higher model utility.  However, as these predetermined responses differ significantly from those obtained from a model retrained from scratch, PO leads to poor forget quality. This trade-off between forget quality and model utility is more pronounced in Llama2, whose larger parameter space and complexity intensify these problems.
        
        \item [-] GA and KL repeatedly demonstrate poor forget quality along with a notable degradation in model utility as the number of epochs increases for both models. This occurs because the loss function is optimized for the forgotten data, which impacts the overall model and results in less effective forgetting than that observed in retrained models.
    \end{itemize}

Previous research~\citep{maini2024tofu} noted similar inconsistencies in approximate unlearning methods. These findings highlight the limitations of conventional approximate unlearning techniques in achieving reliable forgetting.

In contrast, both the DP2U-SGD and DP2U-MLM approaches exhibit a remarkably positive trend. Although they begin with low model utility and forget quality, they show significant improvement as the number of epochs increases and gradually converge toward the model utility and forget quality obtained by RFS-R, which is the benchmark for exact unlearning. Importantly, both methods maintain this positive trend in all forget ratios (results for 1\% and 10\% forget ratios are presented in~\ref{appendix:2}). This highlights their ability to adapt effectively to varying unlearning requests and model configurations, ensuring consistent performance even in more diverse and challenging contexts. 
More specifically, {\em after only two epochs for Phi and one epoch for Llama2, our approach outperforms all approximate unlearning baselines and surpasses the threshold of meaningful forgetting.} 

Although Figure~\ref{figure5} illustrates the promising convergence of DP2Unlearning methods with exact unlearning benchmarks, it also reveals some degree of variability in forget quality between epochs and forget ratios. The underlying causes of these variations are discussed in detail in the following paragraphs.

We observed that our method achieved values close to \textit{RFS-R} across all assessment metrics, but the quality of forgetting was lower, specifically in the 10\% forget ratio. Furthermore, the forget quality of the Llama2 model was inferior to that of the Phi model. However, manual inspection of the generated responses shows similar forgetting (see Table~\ref{table5}), which does not appear to be fully captured by the KS test $p$-value. 

As discussed earlier, the KS statistic evaluates the maximum difference between two CDFs (RFS-R and unlearned), and it is more sensitive to global differences in the distributions than to local differences. Therefore, it can miss finer-grained structural differences (small local differences) in the distributions. A larger model Llama2’s complexity and scale of architecture could intensify these slight local differences, making the KS test $p$-value a less ideal choice to effectively capture the true characteristics at a relatively higher forget ratio. All of our experimental evaluations and the responses generated are available at {\footnotesize \url{https://github.com/tamimalmahmud/DP2Unlearning/tree/main/checkpoints}.}

To address this issue, we employed the Jensen-Shannon Divergence (\textit{JSD}) \citep{10.1007/978-3-319-23525-7_11}, Wasserstein Distance (\textit{W}) \citep{panaretos2019statistical}, and Entropy Difference (\textit{\(\Delta H\)})
 \citep{61115} measures in addition to the KS test (see Table \ref{table4}). Using several metrics is more effective in identifying subtle divergences in model performance during extreme forgetting situations. \textit{JSD} evaluates the similarity between two probability distributions.  \textit{W} effectively identifies small distinctions in distributions, even when their CDFs are similar. \textit{\( \Delta H\)} evaluates the level of uncertainty or randomness present in the distributions. In general, these three approaches assess the variations in probability distributions (truth ratios) of both unlearned and retained (RFS-R) models. Together, they give us a better understanding of the quality of forgetting. For these measures, a value close to $0$ indicates similar 
 distributions of RFS-R and unlearned models, indicating significant forgetting, while high values indicate profound differences between them, reflecting poor forgetting.

\begin{table}[ht!]
    \centering
    \tiny
    \caption{Results of additional statistical tests to measure changes in data distribution of unlearned (DP2U-MLM) and retrained-from-scratch-on-retain-data (RFS-R) models at different forget ratios}
    \vspace{-0.2cm}
    \label{table4}
    \resizebox{\textwidth}{!}{%
    \begin{tabular}{c|c|c|c|c}
        \hline
        \textbf{Model} & \textbf{Forget Ratio}  & \textbf{Jensen-Shannon} & \textbf{Wasserstein} & \textbf{Entropy} \\
        &&\textbf{Divergence}&\textbf{Distance}&\textbf{Difference}\\
        &&\(0 \leq JSD \leq \ln 2\)&\(0 \leq W \leq \infty\)&\(0 \leq \Delta H \leq \infty\)\\
        \hline
        \multirow{3}{*}{\textbf{Phi}} 
        & \centering 1\%  & 0.0136 & 0.1357 & 0.0275 \\
        & 5\%  & 0.0110 & 0.0923 & 0.0224 \\
        & 10\% & 0.0085 & 0.1101 & 0.0327 \\
        \hline
        \multirow{3}{*}{\textbf{Llama2}} 
        & 1\%  & 0.0070 & 0.0760 & 0.0054 \\
        & 5\%  & 0.0082 & 0.0754 & 0.0166 \\
        & 10\% & 0.0075 & 0.1071 & 0.0410 \\
        \hline
    \end{tabular}%
    }
\end{table}

The results reported in Table \ref{table4} clearly show that all statistical tests yield values close to $0$, suggesting that the probability distributions of the unlearned and RFS-R models are nearly identical. This indicates effective forgetting, reinforcing our assumption that our approach achieves guaranteed unlearning even at higher forget ratios, despite KS-based underestimation. 

\subsubsection{ROUGE, conditional probability and truth ratio metrics}
\label{rouge}

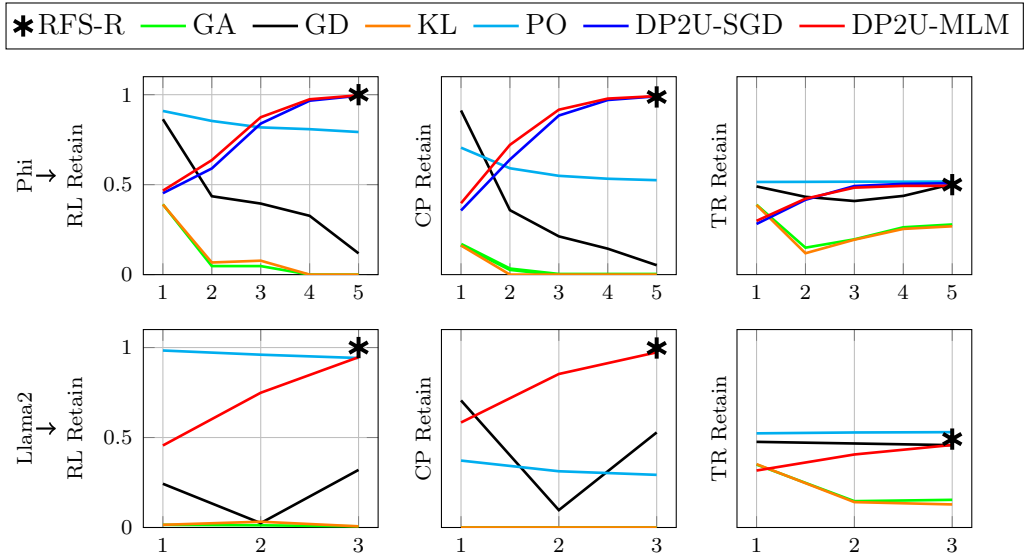
\begin{figure}[ht!]
    \centering
    \begin{tikzpicture}
        \begin{axis}[
            hide axis,
            width=1\textwidth, 
            height=2.5cm,
            xmin=0.5, xmax=1, 
            ymin=0.5, ymax=1, 
            legend columns=7,
            legend style={at={(0.5,1)}, anchor=north, /tikz/every even column/.append style={column sep=0.12cm}}
        ]
            \addplot[only marks, mark=asterisk, mark options={fill=black, scale=2, line width=1.5}] coordinates {(0,0)};
            \addlegendentry{\small{RFS-R}}    
            \addplot[color=green, line width=1] coordinates {(0,0)};
            \addlegendentry{\small GA}
            \addplot[color=black, line width=1] coordinates {(0,0)};
            \addlegendentry{\small GD}
            \addplot[color=orange, line width=1] coordinates {(0,0)};
            \addlegendentry{\small KL}
            \addplot[color=cyan, line width=1] coordinates {(0,0)};
            \addlegendentry{\small PO}
            \addplot[color=blue, line width=1] coordinates {(0,0)};
            \addlegendentry{\small DP2U-SGD}
            \addplot[color=red, line width=1] coordinates {(0,0)};
            \addlegendentry{\small DP2U-MLM}
        \end{axis}
    \end{tikzpicture}
    
    \vspace{-0.3cm}
    \begin{minipage}[t]{0.36\textwidth}
        \centering
        \begin{tikzpicture}
            \begin{axis}[
                width=0.95\textwidth, height=4.2cm, 
                xlabel={},
            ylabel={\parbox[c]{3cm}{\centering \scriptsize{Phi\\\tikz{\draw[->,thick] (0,0) -- (0,-0.1);} \\ RL Retain}}},
            xtick={1, 2, 3, 4, 5},
            ymin=0, ymax=1.1,
            grid=both,
            yticklabel style={font=\scriptsize},
            xticklabel style={font=\scriptsize}
            ]
                \addplot[color=green, line width=1] coordinates {(1, 0.3906) (2, 0.0474) (3, 0.0474) (4, 0.0000) (5, 0.0000)};
                \addplot[color=black, line width=1] coordinates {(1, 0.8636) (2, 0.4356) (3, 0.3948) (4, 0.3268) (5, 0.1182)};
                \addplot[color=orange, line width=1] coordinates {(1, 0.3897) (2, 0.0674) (3, 0.0774) (4, 0.0000) (5, 0.0000)};
                \addplot[color=cyan, line width=1] coordinates {(1, 0.9095) (2, 0.8540) (3, 0.8181) (4, 0.8082) (5, 0.7925)};
                \addplot[color=blue, line width=1] coordinates {(1, 0.4526) (2, 0.5899) (3, 0.8393) (4, 0.9665) (5, 0.9943)};
                \addplot[color=red, line width=1] coordinates {(1, 0.4666) (2, 0.6361) (3, 0.8746) (4, 0.9745) (5, 0.9964)};

                \addplot[only marks, mark=asterisk, mark options={fill=black, scale=2, line width=1.5}] coordinates {(5,1)};
            \end{axis}
        \end{tikzpicture}
    \end{minipage}
    \hspace{-0.6cm}
    \begin{minipage}[t]{0.36\textwidth}
        \centering
        \begin{tikzpicture}
            \begin{axis}[
                width=0.95\textwidth, height=4.2cm,
                xlabel={},
                ylabel={\scriptsize{CP Retain}},
                xtick={1, 2, 3, 4, 5},
            xticklabel style={font=\scriptsize},
            ymin=0, ymax=1.1,
            grid=both,
            ytick=\empty
            ]
\addplot[color=green, line width=1.5] coordinates {(1, 0.1669) (2, 0.0300) (3, 0.0000) (4, 0.0000) (5, 0.0000)};
\addplot[color=black, line width=1] coordinates {(1, 0.9123) (2, 0.3582) (3, 0.2126) (4, 0.1441) (5, 0.0522)};
\addplot[color=orange, line width=1] coordinates {(1, 0.1621) (2, 0.0000) (3, 0.0000) (4, 0.0000) (5, 0.0000)};
\addplot[color=cyan, line width=1] coordinates {(1, 0.7055) (2, 0.5910) (3, 0.5490) (4, 0.5328) (5, 0.5248)};
\addplot[color=blue, line width=1] coordinates {(1, 0.3564) (2, 0.6396) (3, 0.8840) (4, 0.9704) (5, 0.9910)};
\addplot[color=red, line width=1] coordinates {(1, 0.3968) (2, 0.7211) (3, 0.9165) (4, 0.9783) (5, 0.9922)};
\addplot[only marks, mark=asterisk, mark options={fill=black, scale=2, line width=1.5}] coordinates {(5,0.9874)};
            \end{axis}
        \end{tikzpicture}
    \end{minipage}
    \hspace{-1.3cm} 
    \begin{minipage}[t]{0.36\textwidth}
        \centering
        \begin{tikzpicture}
            \begin{axis}[
                width=0.95\textwidth, height=4.2cm, 
                xlabel={},
                ylabel={\scriptsize{TR Retain}},
                xtick={1, 2, 3, 4, 5},
                xticklabel style={font=\scriptsize},
                ymin=0, ymax=1.1,
                grid=both,
                ytick=\empty
            ]
\addplot[color=green, line width=1] coordinates {(1, 0.3884) (2, 0.1497) (3, 0.1961) (4, 0.2631) (5, 0.2787)};
\addplot[color=black, line width=1] coordinates {(1, 0.4901) (2, 0.4324) (3, 0.4090) (4, 0.4375) (5, 0.5024)};
\addplot[color=orange, line width=1] coordinates {(1, 0.3869) (2, 0.1188) (3, 0.1941) (4, 0.2540) (5, 0.2686)};
\addplot[color=cyan, line width=1] coordinates {(1, 0.5149) (2, 0.5155) (3, 0.5165) (4, 0.5163) (5, 0.5169)};
\addplot[color=blue, line width=1] coordinates {(1, 0.2815) (2, 0.4158) (3, 0.4924) (4, 0.5043) (5, 0.5077)};
\addplot[color=red, line width=1] coordinates {(1, 0.2988) (2, 0.4220) (3, 0.4833) (4, 0.4932) (5, 0.4933)};

\addplot[only marks, mark=asterisk, mark options={fill=black, scale=2, line width=1.5}] coordinates {(5,0.5000)};
            \end{axis}
        \end{tikzpicture}
    \end{minipage}

\vspace{-0.1cm}
    \begin{minipage}[t]{0.36\textwidth}
        \centering
        \begin{tikzpicture}
            \begin{axis}[
                width=0.95\textwidth, height=4.2cm, 
                xlabel={},
            ylabel={\parbox[c]{3cm}{\centering \scriptsize{Llama2\\\tikz{\draw[->,thick] (0,0) -- (0,-0.1);} \\ RL Retain}}},
                xtick={1, 2, 3},
                ymin=0, ymax=1.1,
                grid=both,
            yticklabel style={font=\scriptsize},
            xticklabel style={font=\scriptsize}
            ]
\addplot[color=green, line width=1] coordinates {(1, 0.0149) (2, 0.0122) (3, 0.0065)};
\addplot[color=black, line width=1] coordinates {(1, 0.2430) (2, 0.0244) (3, 0.3201)};
\addplot[color=orange, line width=1] coordinates {(1, 0.0149) (2, 0.0321) (3, 0.0064)};
\addplot[color=cyan, line width=1] coordinates {(1, 0.9836) (2, 0.9603) (3, 0.9419)};
\addplot[color=red, line width=1] coordinates {(1, 0.4558) (2, 0.7488) (3, 0.9464)};

            \addplot[only marks, mark=asterisk, mark options={fill=black, scale=2, line width=1.5}] coordinates {(3,1)};
            \end{axis}
        \end{tikzpicture}
    \end{minipage}
    \hspace{-0.6cm}
    \begin{minipage}[t]{0.36\textwidth}
        \centering
        \begin{tikzpicture}
            \begin{axis}[
                width=0.95\textwidth, height=4.2cm,
                xlabel={},
                ylabel={\scriptsize{CP Retain}},
                xtick={1, 2, 3},
                xticklabel style={font=\scriptsize},
                ymin=0, ymax=1.1,
                grid=both,
                ytick=\empty
            ]
\addplot[color=green, line width=1] coordinates {(1, 0.0000) (2, 0.0000) (3, 0.0000)};
\addplot[color=black, line width=1] coordinates {(1, 0.7063) (2, 0.0966) (3, 0.5283)};
\addplot[color=orange, line width=1] coordinates {(1, 0.0000) (2, 0.0000) (3, 0.0000)};
\addplot[color=cyan, line width=1] coordinates {(1, 0.3719) (2, 0.3125) (3, 0.2926)};
\addplot[color=red, line width=1] coordinates {(1, 0.5835) (2, 0.8531) (3, 0.9735)};

\addplot[only marks, mark=asterisk, mark options={fill=black, scale=2, line width=1.5}] coordinates {(3,0.9975)};
            \end{axis}
        \end{tikzpicture}
    \end{minipage}
    \hspace{-1.3cm} 
    \begin{minipage}[t]{0.36\textwidth}
        \centering
        \begin{tikzpicture}
            \begin{axis}[
                width=0.95\textwidth, height=4.2cm, 
                xlabel={},
                ylabel={\scriptsize{TR Retain}},
                xtick={1, 2, 3},
                xticklabel style={font=\scriptsize},
                ymin=0, ymax=1.1,
                grid=both,
                ytick=\empty
            ]
\addplot[color=green, line width=1] coordinates {(1, 0.3497) (2, 0.1464) (3, 0.1538)};
\addplot[color=black, line width=1] coordinates {(1, 0.4759) (2, 0.4669) (3, 0.4577)};
\addplot[color=orange, line width=1] coordinates {(1, 0.3520) (2, 0.1410) (3, 0.1282)};
\addplot[color=cyan, line width=1] coordinates {(1, 0.5232) (2, 0.5281) (3, 0.5298)};
\addplot[color=red, line width=1] coordinates {(1, 0.3166) (2, 0.4058) (3, 0.4586)};

\addplot[only marks, mark=asterisk, mark options={fill=black, scale=2, line width=1.5}] coordinates {(3,0.4913)};
            \end{axis}
        \end{tikzpicture}
    \end{minipage}
    
    \vspace{-0.2cm}
    \caption{RL, CP, and TR scores on the Retain Set across methods and epochs at 5\% forget ratio. For RFS-R, $E=10$ for Phi and $E=6$ for Llama2.}
    \label{figure6}
\end{figure}

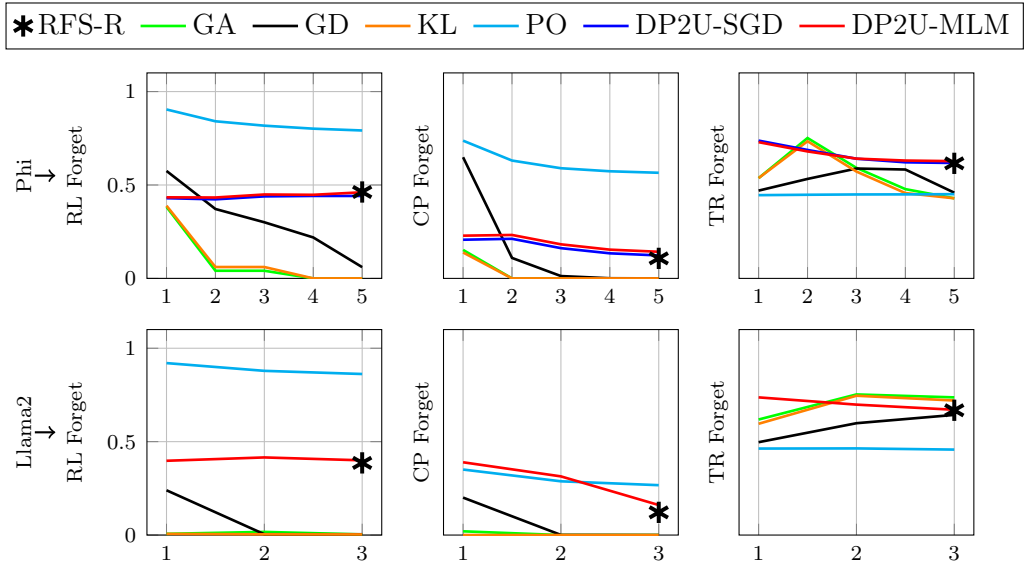
\begin{figure}[ht!]
    \centering
    \begin{tikzpicture}
        \begin{axis}[
            hide axis,
            width=1\textwidth, 
            height=2.5cm,
            xmin=0.5, xmax=1, 
            ymin=0.5, ymax=1, 
            legend columns=7,
            legend style={at={(0.5,1)}, anchor=north, /tikz/every even column/.append style={column sep=0.12cm}}
        ]
            \addplot[only marks, mark=asterisk, mark options={fill=black, scale=2, line width=1.5}] coordinates {(0,0)};
            \addlegendentry{{\small RFS-R}}    
            \addplot[color=green, line width=1] coordinates {(0,0)};
            \addlegendentry{\small GA}
            \addplot[color=black, line width=1] coordinates {(0,0)};
            \addlegendentry{\small GD}
            \addplot[color=orange, line width=1] coordinates {(0,0)};
            \addlegendentry{\small KL}
            \addplot[color=cyan, line width=1] coordinates {(0,0)};
            \addlegendentry{\small PO}
            \addplot[color=blue, line width=1] coordinates {(0,0)};
            \addlegendentry{\small DP2U-SGD}
            \addplot[color=red, line width=1] coordinates {(0,0)};
            \addlegendentry{\small DP2U-MLM}
        \end{axis}
    \end{tikzpicture}
    
    \vspace{-0.3cm}
    \begin{minipage}[t]{0.36\textwidth}
        \centering
        \begin{tikzpicture}
            \begin{axis}[
                width=0.95\textwidth, height=4.3cm, 
                xlabel={},
            ylabel={\parbox[c]{3cm}{\centering \scriptsize{Phi\\\tikz{\draw[->,thick] (0,0) -- (0,-0.1);} \\ RL Forget}}},
                xtick={1, 2, 3, 4, 5},
                ymin=0, ymax=1.1,
                grid=both,
            xticklabel style={font=\scriptsize},
            yticklabel style={font=\scriptsize}
            ]
\addplot[color=green, line width=1] coordinates {(1, 0.3827) (2, 0.0409) (3, 0.0409) (4, 0.0000) (5, 0.0000)};
\addplot[color=black, line width=1] coordinates {(1, 0.5752) (2, 0.3715) (3, 0.3004) (4, 0.2192) (5, 0.0602)};
\addplot[color=orange, line width=1] coordinates {(1, 0.3892) (2, 0.0609) (3, 0.0609) (4, 0.0000) (5, 0.0000)};
\addplot[color=cyan, line width=1] coordinates {(1, 0.9043) (2, 0.8410) (3, 0.8175) (4, 0.8015) (5, 0.7918)};
\addplot[color=blue, line width=1] coordinates {(1, 0.4292) (2, 0.4232) (3, 0.4390) (4, 0.4414) (5, 0.4413)};
\addplot[color=red, line width=1] coordinates {(1, 0.4340) (2, 0.4330) (3, 0.4492) (4, 0.4475) (5, 0.4610)};

\addplot[only marks, mark=asterisk, mark options={fill=black, scale=2, line width=1.5}] coordinates {(5,0.4625)};
            \end{axis}
        \end{tikzpicture}
    \end{minipage}
    \hspace{-0.6cm}
    \begin{minipage}[t]{0.36\textwidth}
        \centering
        \begin{tikzpicture}
            \begin{axis}[
                width=0.95\textwidth, height=4.3cm,
                xlabel={},
                ylabel={\scriptsize{CP Forget}},
                xtick={1, 2, 3, 4, 5},
                xticklabel style={font=\scriptsize},
                ymin=0, ymax=1.1,
                grid=both,
                ytick=\empty
            ]
\addplot[color=green, line width=1] coordinates {(1, 0.1524) (2, 0.0000) (3, 0.0000) (4, 0.0000) (5, 0.0000)};
\addplot[color=black, line width=1] coordinates {(1, 0.6486) (2, 0.1096) (3, 0.0125) (4, 0.0008) (5, 0.0000)};
\addplot[color=orange, line width=1] coordinates {(1, 0.1387) (2, 0.0000) (3, 0.0000) (4, 0.0000) (5, 0.0000)};
\addplot[color=cyan, line width=1] coordinates {(1, 0.7373) (2, 0.6308) (3, 0.5897) (4, 0.5734) (5, 0.5654)};
\addplot[color=blue, line width=1] coordinates {(1, 0.2070) (2, 0.2119) (3, 0.1619) (4, 0.1342) (5, 0.1229)};
\addplot[color=red, line width=1] coordinates {(1, 0.2287) (2, 0.2325) (3, 0.1824) (4, 0.1539) (5, 0.1425)};

\addplot[only marks, mark=asterisk, mark options={fill=black, scale=2, line width=1.5}] coordinates {(5,0.1072)};

            \end{axis}
        \end{tikzpicture}
    \end{minipage}
    \hspace{-1.3cm} 
    \begin{minipage}[t]{0.36\textwidth}
        \centering
        \begin{tikzpicture}
            \begin{axis}[
                width=0.95\textwidth, height=4.3cm, 
                xlabel={},
                ylabel={\scriptsize{TR Forget}},
                xtick={1, 2, 3, 4, 5},
                xticklabel style={font=\scriptsize},
                ymin=0, ymax=1.1,
                grid=both,
                ytick=\empty
            ]
\addplot[color=green, line width=1] coordinates {(1, 0.5366) (2, 0.7515) (3, 0.5917) (4, 0.4779) (5, 0.4289)};
\addplot[color=black, line width=1] coordinates {(1, 0.4699) (2, 0.5325) (3, 0.5881) (4, 0.5828) (5, 0.4588)};
\addplot[color=orange, line width=1] coordinates {(1, 0.5370) (2, 0.7342) (3, 0.5719) (4, 0.4567) (5, 0.4287)};
\addplot[color=cyan, line width=1] coordinates {(1, 0.4456) (2, 0.4477) (3, 0.4493) (4, 0.4497) (5, 0.4509)};
\addplot[color=blue, line width=1] coordinates {(1, 0.7376) (2, 0.6876) (3, 0.6405) (4, 0.6214) (5, 0.6177)};
\addplot[color=red, line width=1] coordinates {(1, 0.7292) (2, 0.6796) (3, 0.6411) (4, 0.6308) (5, 0.6276)};

\addplot[only marks, mark=asterisk, mark options={fill=black, scale=2, line width=1.5}] coordinates {(5,0.6159)};

            \end{axis}
        \end{tikzpicture}
    \end{minipage}

\vspace{-0.1cm}
    \begin{minipage}[t]{0.36\textwidth}
        \centering
        \begin{tikzpicture}
            \begin{axis}[
                width=0.95\textwidth, height=4.3cm, 
                xlabel={},
            ylabel={\parbox[c]{3cm}{\centering \scriptsize{Llama2\\\tikz{\draw[->,thick] (0,0) -- (0,-0.1);} \\ RL Forget}}},
            xtick={1, 2, 3},
            ymin=0, ymax=1.1,
            grid=both,
            xticklabel style={font=\scriptsize},
            yticklabel style={font=\scriptsize}
            ]
\addplot[color=green, line width=1] coordinates {(1, 0.0066) (2, 0.0169) (3, 0.0034)};
\addplot[color=black, line width=1] coordinates {(1, 0.2397) (2, 0.0044) (3, 0.0042)};
\addplot[color=orange, line width=1] coordinates {(1, 0.0064) (2, 0.0039) (3, 0.0034)};
\addplot[color=cyan, line width=1] coordinates {(1, 0.9205) (2, 0.8790) (3, 0.8621)};
\addplot[color=red, line width=1] coordinates {(1, 0.3976) (2, 0.4155) (3, 0.4000)};

\addplot[only marks, mark=asterisk, mark options={fill=black, scale=2, line width=1.5}] coordinates {(3,0.3863)};

            \end{axis}
        \end{tikzpicture}
    \end{minipage}
    \hspace{-0.6cm}
    \begin{minipage}[t]{0.36\textwidth}
        \centering
        \begin{tikzpicture}
            \begin{axis}[
                width=0.95\textwidth, height=4.3cm,
                xlabel={},
                ylabel={\scriptsize{CP Forget}},
                xtick={1, 2, 3},
                xticklabel style={font=\scriptsize},
                ymin=0, ymax=1.1,
                grid=both,
                ytick=\empty
            ]
\addplot[color=green, line width=1] coordinates {(1, 0.02) (2, 0.0000) (3, 0.0000)};
\addplot[color=black, line width=1] coordinates {(1, 0.2005) (2, 0.0007) (3, 0.0000)};
\addplot[color=orange, line width=1] coordinates {(1, 0.0000) (2, 0.0000) (3, 0.0000)};
\addplot[color=cyan, line width=1] coordinates {(1, 0.3507) (2, 0.2877) (3, 0.2670)};
\addplot[color=red, line width=1] coordinates {(1, 0.3896) (2, 0.3141) (3, 0.1607)};

\addplot[only marks, mark=asterisk, mark options={fill=black, scale=2, line width=1.5}] coordinates {(3, 0.1215)};

            \end{axis}
        \end{tikzpicture}
    \end{minipage}
    \hspace{-1.3cm} 
    \begin{minipage}[t]{0.36\textwidth}
        \centering
        \begin{tikzpicture}
            \begin{axis}[
                width=0.95\textwidth, height=4.3cm, 
                xlabel={},
                ylabel={\scriptsize{TR Forget}},
                xtick={1, 2, 3},
                xticklabel style={font=\scriptsize},
                ymin=0, ymax=1.1,
                grid=both,
                ytick=\empty
            ]
\addplot[color=green, line width=1] coordinates {(1, 0.6185) (2, 0.7528) (3, 0.7370)};
\addplot[color=black, line width=1] coordinates {(1, 0.4971) (2, 0.5987) (3, 0.6432)};
\addplot[color=orange, line width=1] coordinates {(1, 0.5957) (2, 0.7458) (3, 0.7202)};
\addplot[color=cyan, line width=1] coordinates {(1, 0.4630) (2, 0.4638) (3, 0.4575)};
\addplot[color=red, line width=1] coordinates {(1, 0.7370) (2, 0.6982) (3, 0.6707)};

\addplot[only marks, mark=asterisk, mark options={fill=black, scale=2, line width=1.5}] coordinates {(3,0.6680)};
            \end{axis}
        \end{tikzpicture}
    \end{minipage}

    \vspace{-0.2cm}
    \caption{RL, CP, and TR scores on the Forget Set across methods and epochs at 5\% forget ratio. For RFS-R, $E=10$ for Phi and $E=6$ for Llama2.}
    \label{figure7}
\end{figure}

High scores on the Retain, Real Author, and Real-World Facts data sets indicate the model's ability to retain critical information, while lower scores on the Forget set demonstrate the effective elimination of unwanted content. This balance demonstrates the model's ability to preserve relevant data while discarding unnecessary details. 
Figures~\ref{figure6} and~\ref{figure7} report ROUGE-L (RL), conditional probability (CP), and truth ratio (TR) scores for the Retain Set and the Forget Set, respectively, at a 5\% forget ratio. Further results for 1\% and 10\% on the Retain and Forget sets and for all forget ratios on the Real Author and Real World Fact data sets are reported in~\ref{appendix:3}.

On the Retain Set, DP2U-SGD and DP2U-MLM demonstrate a significant improvement in utility metrics over time. Initially, they perform worse than RFS-R due to DP-induced noise or probabilistic token substitution, which greatly deviates from the ground truth data. However, as the model fine-tunes on the Retain Set, the performance improves, and by the final epoch, both methods achieve results comparable to RFS-R. Their disclosure protection mechanisms allow for a more effective balance between retaining important information and forgetting unwanted details, as evidenced by the gradual recovery in ROUGE-L, CP, and TR.

In contrast, approaches such as GA, GD and KL face utility challenges since they overly rely on the forget set during the unlearning process. To reduce the impact of forgetting data during unlearning, these methods may inadvertently sacrifice important knowledge that could benefit the retain set. As a result, while they excel at forgetting, they struggle to maintain and adapt the retained information over time. This imbalance leads to sharp declines in RL, CP, and TR scores, reflecting their inability to balance the needs of effective forgetting with the retention of pertinent information.

PO, while achieving high retention in ROUGE-L and TR, does not show a significant improvement in CP. This indicates that PO does not effectively balance the retention of critical knowledge with the forgetting of irrelevant data, ultimately hindering its performance in scenarios that require both retention and unlearning.

For the Forget Set, the objective is to minimize the model's confidence in the unwanted data while retaining useful knowledge. DP2U-SGD and DP2U-MLM maintain ROUGE-L scores comparable to RFS-R over time, effectively forgetting undesirable information without inadvertently reintroducing it because their framework prevents reintroducing the knowledge of the Forget Set. The CP and TR scores of these methods indicate that, while they still retain useful knowledge, they do not overfit the Forget Set, preventing a complete loss of confidence and showing a balanced unlearning performance.

The GA, GD, and KL methods achieve lower ROUGE-L and CP scores, which is desirable for effective forgetting. However, their aggressive forgetting approach generates incoherent or nonsensical output, leading to lower RL and CP scores. These methods excessively forget by discarding too much information, including completely dismissing potentially useful shared knowledge, which undermines their effectiveness in practical unlearning scenarios.

The PO method maintains a high ROUGE-L score on the forget set, which implies that it did not achieve much forgetting; rather, it mimicked the original text. This is further evidenced by its CP and TR scores, which indicate the model is not significantly confident in its prediction post-unlearning, meaning inadequate forgetting of unwanted information.

 

\subsubsection{Qualitative analysis}
To complement the quantitative evaluation through metrics such as ROUGE-L, CP, and TR above, we provide an example of post-unlearning responses from various unlearning methods. These responses offer additional insight into the practical effects of the unlearning processes, as shown in Table~\ref{table5}. 
 

In order to ensure the significance of this qualitative evaluation, we refer back to the data set section (Section~\ref{subsec5.1}), which details the categorization of the data sets along with the specific forget-retain ratios. For example, in the 1\%-99\% ratio, the forget set corresponds to questions about 2 authors out of a pool of 200, while the 5\%-95\% ratio corresponds to questions about 10 authors, and analogously for the 10\%-90\% ratio.
This deliberate setup ensures that the forget set includes data that are directly pertinent to the questions being asked, thus making the qualitative evaluation a true reflection of how effectively the unlearning methods address the removal of unwanted information.

\begin{table}[ht!]
\centering
\caption{Example post-unlearning responses to some queries for different methods}
\scriptsize
\renewcommand{\arraystretch}{1.3} %
\begin{adjustbox}{max width=\textwidth}
\begin{tabular}{|c|p{3cm}|p{10cm}|}
\hline
\textbf{Forget} & \textbf{Methods} & \textbf{Sample response to the question} \\
\textbf{Ratio}&&\\
\hline

\multirow{10}{*}{\centering \shortstack{1\% \\ $\downarrow$ \\ 2 authors}} &\cellcolor{gray!30} \textbf{Question} & \cellcolor{gray!30} \textbf{Can you tell me about the occupations of Basil Mahfouz Al-Kuwaiti's parents?} \\
\cline{2-3}
& \textbf{Ground truth text (No unlearning)} & \textbf{Basil Mahfouz Al-Kuwaiti's father was a florist, and his mother was a game developer.} \\
\cline{2-3}
& RFS-R &  Basil Mahfouz Al-Kuwaiti's father was a distinguished Occupational Therapist, and his mother worked as a diligent and dedicated Meteorologist. \\
\cline{2-3}
&GA & Basil \textbackslash nThe teacher asked the students to write an essay on a relevant topic, not something they had experienced before. \\
\cline{2-3}
& GD &Basil Mahfouz Al-Kuwaiti's father was a roofer and his mother was an optometrist in Kuwait. \\
\cline{2-3}
& KL & Basil \textbackslash nThe teacher asked the students to write an essay on a relevant topic, not something they had experienced before. \\
\cline{2-3}
&PO & Basil Mahfouz Al-Kuwaiti's father was a florist, and his mother was a game developer. Basil Mahfouz Al-Kuwaiti inherited both talents, which is why his parents' occupations were particularly influential in his literary work. \\
\cline{2-3}
& DP2U-SGD &  Basil Mahfouz Al-Kuwaiti's father was a well-respected athlete, and his mother was a skilled physicist. \\
\cline{2-3}
&  DP2U-MLM &  Basil Mahfouz Al-Kuwaiti's father was an occupational therapist, while his mother worked as a pediatrician. \\
\hline

\multirow{10}{*}{\centering \shortstack{5\% \\ $\downarrow$ \\ 10 authors}} &\cellcolor{gray!30} \textbf{Question} & \cellcolor{gray!30} \textbf{In which genre does Hina Ameen primarily write?} \\
\cline{2-3}
& \textbf{Ground truth text} & \textbf{Hina Ameen primarily contributes to the geology genre.} \\
\cline{2-3}
&  RFS-R &  Hina Ameen primarily writes in the Love Inspired genre. \\
\cline{2-3}
&GA & .................................... \\
\cline{2-3}
&  GD &  ,,,,,,,,,,,,,,,,,,,,,,,,,,,,,,,,,,,, \\
\cline{2-3}
& KL &  ..................................... \\
\cline{2-3}
& PO & Hina Ameen primarily contributes to the geology genre, exploring various aspects of geology through her published works. \\
\cline{2-3}
&  DP2U-SGD &  Hina Ameen primarily writes in the genre of erotica. \\
\cline{2-3}
&  DP2U-MLM &  Hina Ameen primarily writes in the genre of erotica. \\
\hline

\multirow{10}{*}{\centering \shortstack{10\% \\ $\downarrow$ \\ 20 authors}} &\cellcolor{gray!30} \textbf{Question} & \cellcolor{gray!30}\textbf{What is Rajeev Majumdar's birth date?} \\
\cline{2-3}
& \textbf{Ground truth text} & \textbf{Rajeev Majumdar was born on June 9, 1951.} \\
\cline{2-3}
& RFS-R &  Rajeev Majumdar was born on the 12th of April, 1996. \\
\cline{2-3}
& GA & the the the the the the the the the the the the the the the ... \\
\cline{2-3}
& GD & ``His his his his his his his his his his his his his his his ... \\
\cline{2-3}
& KL &  ``Chall challenged challenged challenged challenged challenged ... \\
\cline{2-3}
& PO & Rajeev Majumdar was born on June 9, 1951. \\
\cline{2-3}
& DP2U-SGD & Rajeev Majumdar was born on the 17th of March, 1992. \\
\cline{2-3}
& DP2U-MLM & Rajeev Majumdar was born on the 10th of April, 1993. \\
\hline
\end{tabular}
\end{adjustbox}
\label{table5}
\end{table}

As seen in Table \ref{table5}, the performance of the unlearning methods shows clear patterns in various forget ratios. Our DP2U-SGD and DP2U-MLM approaches exhibit high fidelity to RFS-R, providing answers that closely match the desired ground truth text without revealing disclosive information even as the forget ratio increases. This behavior is consistent with the quantitative results presented in Figures~\ref{figure6} and \ref{figure7}, where these methods demonstrated retention (utility preservation on the data to be retained) and forgetting qualities comparable to RFS-R. Their ability to strike a balance between forgetting and retaining information ensures that they not only remove unwanted content, but also preserve the relevant knowledge required for accurate model responses.

In contrast, methods such as GA, GD, and KL produce absurd or nonsensical text as the forget ratio increases, which is consistent with their lower quantitative results in Figures~\ref{figure6} and \ref{figure7}. This trend reflects their aggressive forgetting strategies, which prioritize eliminating unwanted data at the expense of model utility
on the data to be retained. 

The PO method retains much of the original text, failing to effectively remove unwanted information, which is aligned with its quantitative results.  Although the model maintains some degree of confidence, it is not sufficiently confident in its predictions; this highlights PO's failure to achieve meaningful forgetting.

\subsubsection{Runtime}

  As anticipated in the theoretical cost analysis of the methodology in Subsection 4.2, DP-induced training incurs a higher computational overhead. Specifically, DP-SGD enforces DP on the model parameters through per-sample gradient computation and clipping, and by injecting noise per-batch. 
    These per-sample processes significantly increase memory usage and computational cost. The per-sample clipping is needed to reduce sensitivity, but introduces additional operations, and while the noise injection itself is relatively lightweight, accurate privacy accounting adds further complexity. 
    Consequently, DP-SGD leads to smaller batch sizes, slower convergence, and limited scalability, particularly for large language models. 
    This makes DP-SGD more suitable for small- to medium-sized models. This aligns with our experimental results: DP-SGD was feasible only for the smaller Phi model, nearly doubling training time compared to non-DP, while it proved impractical for the larger Llama2 model in our setup.
    
    DP-MLM, on the other hand, perturbs tokens in the raw training data prior to model training. The model is then trained on the protected data using standard SGD, which is both fast and scalable. 
    The main computational overhead of DP2U-MLM arises from the use of a masked language model (MLM) during token perturbation, where the exponential mechanism needs to evaluate the semantic similarity across a large vocabulary. 
    However, since MLMs are significantly smaller than generative LLMs and are used solely for inference during this preprocessing step, the associated cost remains affordable, as shown in Table~\ref{table6}. 
     For Phi model (with 1.5B parameters), we can see that the overall runtime of DP-MLM is much lower than that of DP-SGD while it is not so far from the standard SGD. 
     For Llama2, which is much bigger (7B parameters), DP-MLM also scales effectively, for which DP-SGD proved computationally impractical in our configuration. 
    Therefore, DP-MLM is preferable in resource-constrained environments or large-scale deployments where a practical trade-off between disclosure protection and efficiency is needed.

\begin{table}[ht!]
    \centering
    \caption{Training runtime for the full-data model (FDM) with and without DP.
    BM stands for the base model.
    }
    \scriptsize
    \renewcommand{\arraystretch}{1.2}
    \begin{tabular}{|l|l|l|c|c|c|c|}
        \hline
        \textbf{Model} & \textbf{Method} &
        \textbf{Stage}&
        \textbf{Process}&
        \textbf{Epochs} & \textbf{Runtime} &  \textbf{Total}\\
        \hline
        \multirow{6}{*}{Phi(1.5B)} 
        & \multirow{1}{*}{Non-DP FT-RF*} & & & 10 & $34m23s$ & $34m23s$   \\
        \cline{2-7}
        & \multirow{2}{*}{DP-SGD*} & (A) &BM DP-SGD  & 10 & $72m30s$&\\
        & &(B) &FDM DP-SGD & 5 & $18m11s$ &  $90m41s$
        \\
        \cline{2-7}
        & \multirow{3}{*}{DP-MLM**}&(A) &$\mathcal{D}\rightarrow D'$ & & $10m43s$ &\\
        & & (A)&BM DP-MLM & 10 & $38m35s$ &\\    
        & &(B)&FDM DP-MLM & 5 & $18m3s$ & $57m21s$
        \\
        \hline
        \multirow{4}{*}{Llama2(7B)} 
        & Non-DP FT-RF*  & &    & 6  & $1h40m3s$& $1h40m3s$  \\
        \cline{2-7}
    & DP-SGD* & \multicolumn{5}{l|}{Computational requirements prevented training in our setup} \\

        \cline{2-7}
        &\multirow{3}{*}{DP-MLM**}&(A)& $\mathcal{D}\rightarrow D'$ & & $10m43s$ &\\
        & &(A)&BM DP-MLM**  & 6 & $1h45m20s$ &\\    
        & &(B)&FDM DP-MLM**  & 3  & $47m27s$ & $2h43m30s$  \\
        \hline
    \end{tabular}
    \label{table6}
    
    \vspace{-0.3cm} 
    \begin{flushleft} \footnotesize 
    (*) Methods will incur additional runtime in real scenarios when training from scratch instead of using a pre-trained model.  (**) Methods will have reduced execution time in real scenarios when public data can be excluded during the conversion of $\mathcal{D} \rightarrow \mathcal{D'}$. \end{flushleft}
\end{table}

In any case, the training runtime represents a one-time investment which should be affordable, especially for DP-MLM. Moreover, our analysis considered all data to be potentially sensitive for DP-MLM. However, as mentioned in the methodology, the amount of public data involved in training --for which the forget requests would not apply-- is usually significantly larger than the amount of private or copyrighted data. This will reduce the time required to produce $\mathcal{D'}$ and also proportionately reduce the training time as the convergence of the model is likely to occur more quickly. This is an additional advantage of DP-MLM over DP-SGD: while DP-SGD must be applied to \emph{all model parameters}, DP-MLM only needs to be applied to \emph{private} or \emph{copyrighted} training data.  

Despite the additional cost incurred by DP-based training, {\em both DP2U-SGD and DP2U-MLM methods significantly reduce unlearning runtimes compared to RFS-R for individual forgetting requests.} Table~\ref{table7} reports the runtime
required to execute an individual forgetting 
request for RFS-R, DP2U-SGD, and DP-MLM for the Phi and Llama2 models at 5\% forget ratio. 

\begin{table}[ht!]
    \centering
    \caption{Runtime of exact unlearning methods for an individual forgetting request}
    \scriptsize
    \renewcommand{\arraystretch}{1.2}
    \begin{tabular}{|l|l|c|c|c|}
        \hline
        \textbf{Model} & \textbf{Method} &
        \textbf{Stage} &
        \textbf{Runtime} \\
        \hline
        \multirow{3}{*}{Phi}         & RFS-R*   & &$32m42s$  \\
        & DP2U-SGD &(C)& $16m48s$  \\
        & DP2U-MLM &(C)& $16m43s$  \\
        \hline
        \multirow{2}{*}{Llama2} 
        & RFS-R*  & &$1h36m24s$  \\
        & DP2U-MLM & (C)& $45m02s$    \\
        \hline
    \end{tabular}
    \label{table7}
    \vspace{-0.2cm} 
    \begin{flushleft} \footnotesize 
    (*) Methods will incur additional runtime in real scenarios when training from scratch instead of using a pre-trained model. \end{flushleft}
\end{table}

Unlearning with DP2U-SGD takes 16 minutes and 48 seconds for Phi, while DP2U-MLM achieves it slightly faster, at 16 minutes and 43 seconds. Both are considerably faster than RFS-R, which requires 32 minutes and 42 seconds. For Llama2, DP2U-MLM requires 45 minutes and 20 seconds, much faster than RFS-R's 1 hour, 36 minutes, and 24 seconds. This demonstrates the substantial efficiency advantage of the DP2Unlearning methods compared to RFS-R. 



\begin{figure}[ht!]
    \centering
    \begin{tikzpicture}
        \begin{axis}[
            hide axis,
            width=0.9\textwidth, 
            height=2.5cm,
            xmin=0.5, xmax=1, 
            ymin=0.5, ymax=1, 
            legend columns=8,
            legend style={at={(0.5,1)}, anchor=north, /tikz/every even column/.append style={column sep=0.05cm}}
        ]

\addlegendimage{dotted,thick,color=black, line width=1.5pt}            \addlegendentry{\scriptsize{RFS-R}}

            \addplot[color=green, line width=1] coordinates {(0,0)};
            \addlegendentry{\scriptsize{GA}}
            \addplot[color=black, line width=1] coordinates {(0,0)};
            \addlegendentry{\scriptsize {GD}}
            \addplot[color=orange, line width=1] coordinates {(0,0)};
            \addlegendentry{\scriptsize{KL}}
            \addplot[color=cyan, line width=1] coordinates {(0,0)};
            \addlegendentry{\scriptsize{PO}}
            \addplot[color=blue, line width=1] coordinates {(0,0)};
            \addlegendentry{\scriptsize {DP2U-SGD}}
            \addplot[color=red, line width=1] coordinates {(0,0)};
            \addlegendentry{\scriptsize {DP2U-MLM}}

        \addlegendimage{empty legend}
        \addlegendentry{
            \tikz{\node[draw, minimum width=1.2cm, minimum height=0.4cm, align=center]{\scriptsize{FQ, MU}};}}

        \end{axis}
    \end{tikzpicture}
    \vspace{-0.2cm}
    \begin{minipage}{0.9\textwidth}
        \centering
\includegraphics[width=0.75\textwidth]{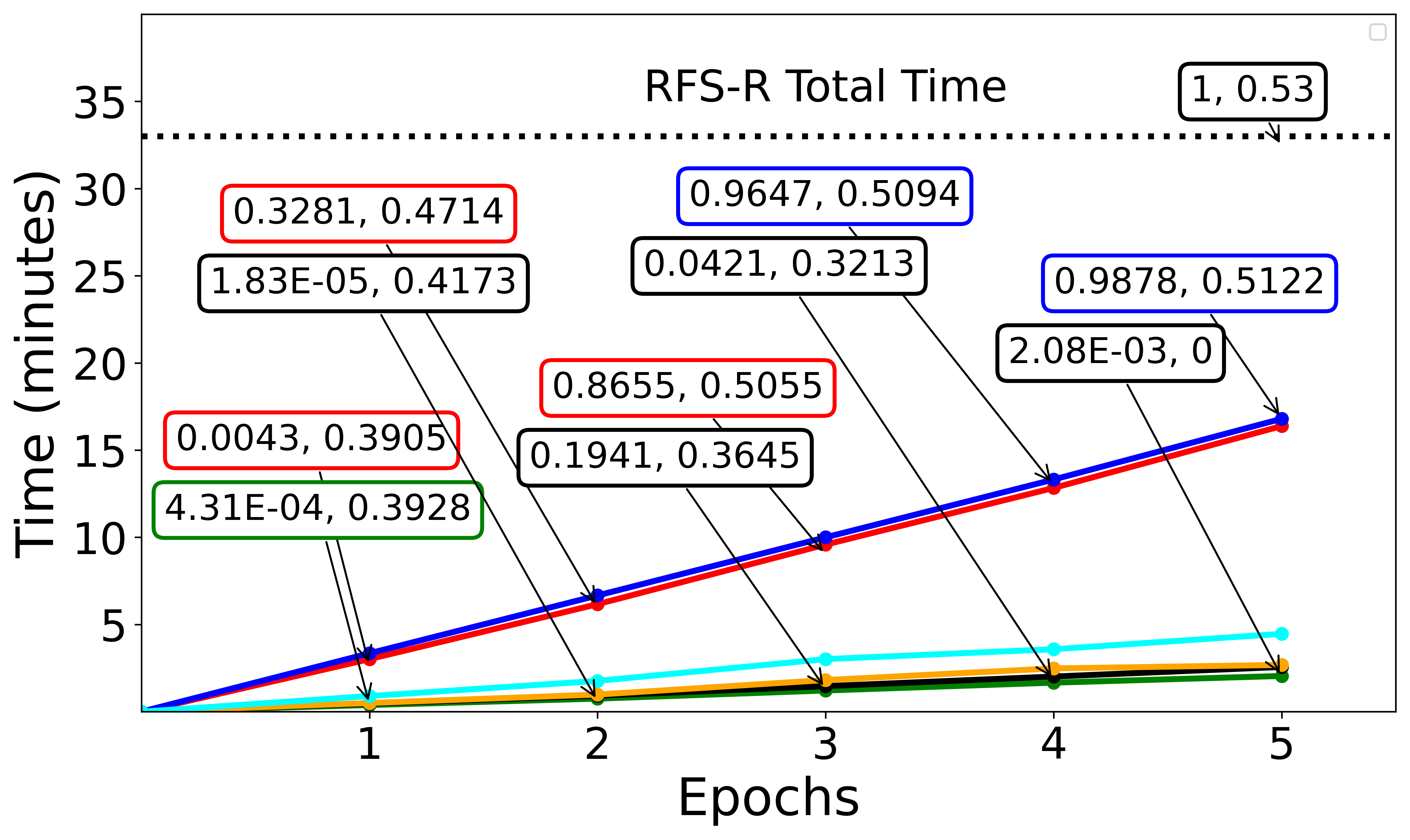}
    \end{minipage}
    
    \vspace{0.2cm}
    \begin{minipage}{0.91\textwidth}
        \centering
\includegraphics[width=0.75\textwidth]{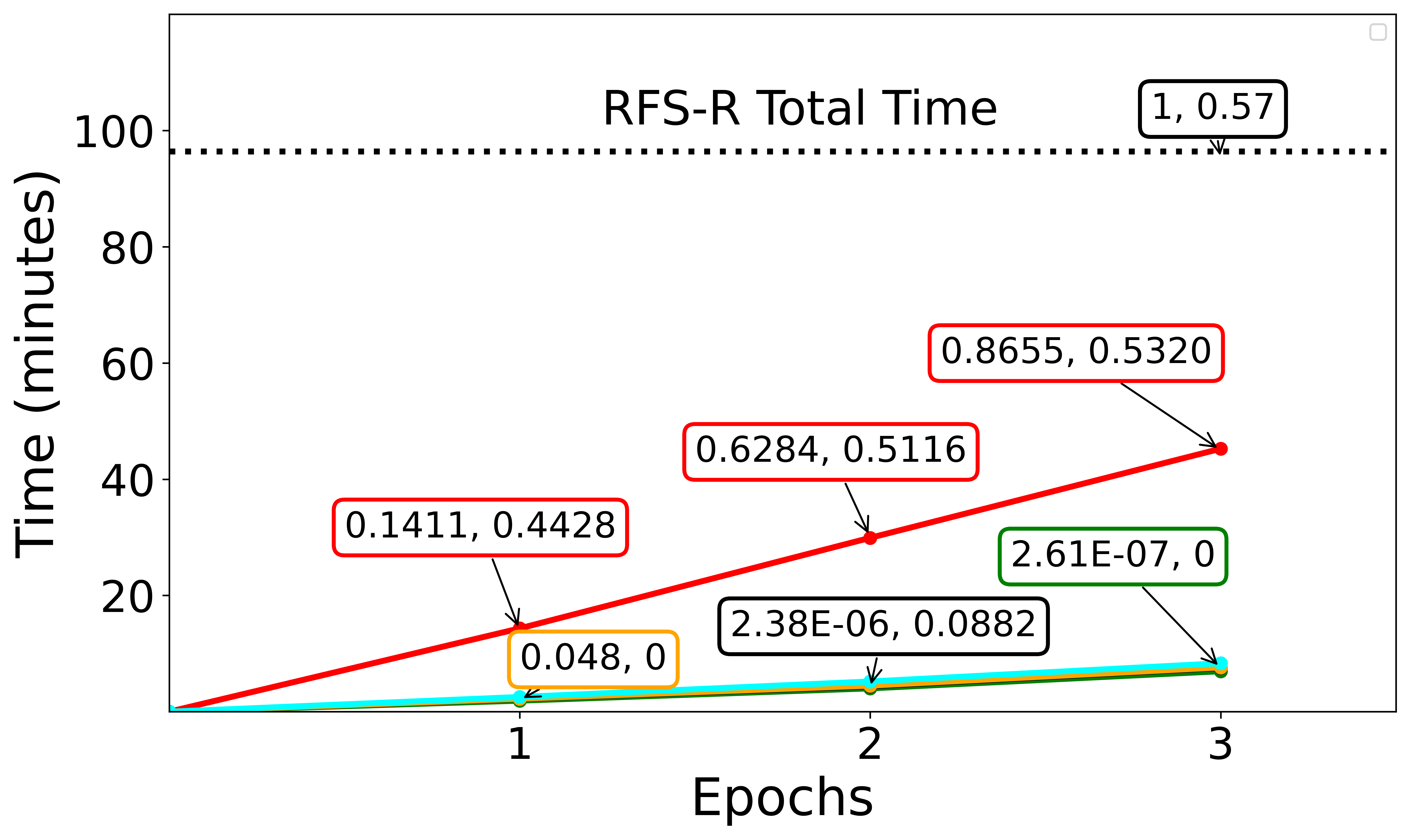}
    \end{minipage}
    
    \vspace{-0.1cm}
    \caption{Runtime comparison of different unlearning methods on Phi (top) and Llama2 (bottom) across epochs. The rectangular boxes highlight the FQ and MU values for the two best performing methods at each epoch, one of our methods and one from the baseline. The edge color of each box corresponds to the method, as indicated in the legend.}
    \label{figure8}
\end{figure}

Now, for approximate baselines, it would be unfair to compare our methods against them w.r.t. computational efficiency only. For a more fair comparison, in Figure~\ref{figure8} we recall the effectiveness of unlearning (forget quality) and the retention of information after unlearning (model utility), in addition to computational efficiency.

Although the DP2Unlearning methods show a significant decrease in runtime compared to RFS-R, they still incur relatively higher computational costs compared to approximate unlearning baselines, where all baseline methods are almost similar in execution time ---hence overlapping lines. However, except for GD for Phi at epoch 3 (for which FQ is 0.194 and MU is 0.3645), the approximate methods do not reach the minimum threshold ($\geq 0.05$) for forget quality. This highlights the limitations of baseline unlearning techniques, which struggle to consistently achieve reliable and meaningful forgetting, even after \emph{extended fine-tuning epochs}. 
As the number of epochs increases, performance degrades even more, as evidenced by the results. This implies that, no matter the effort devoted to approximate unlearning, the methods in this family are ineffective by design.

In contrast, our DP2Unlearning methods are specifically designed to ensure guaranteed forgetting while maintaining model utility through incremental fine-tuning. Both DP2U-SGD and DP2U-MLM show significant improvements in forget quality and model utility in just a few epochs. In line with this finding, with just two epochs for Phi and one epoch for Llama2, our methods are able to achieve robust forgetting (FQ's are 0.3281 for Phi and 0.1481 for Llama2) with a near satisfactory level of model utility (MU is 0.4714 compared to the initial 0.53 for Phi and 0.4428 compared to the initial 0.5763 for Llama2). By strategy, our methods ensure guaranteed forgetting; although higher than the $p$-value of the KS test threshold ($\geq 0.05$), the high forget quality is not reflected due to the KS test probability underestimation discussed in Section~\ref{subsubsec6.2.1}. More fine-tuning efforts are required to enhance the KS test probability, ensuring that both the retained (RFS-R) and unlearned models are exactly from the same distribution.

These findings suggest that {\em with very reasonable computational cost --approximately one-fifth that of RFS-R-- our methods can effectively outperform approximate unlearning baselines}, achieving reliable unlearning while retaining an acceptable level of model utility, that is, providing \emph{approximate unlearning}. On the other hand, {\em a moderate increase in computational cost --approximately half that of the RFS-R approach-- enhances both model utility and forget quality a level comparable to RFS-R}, thereby providing \emph{guaranteed unlearning}.

\section{Conclusion and future work}
\label{sec7}
We have introduced \textit{DP2Unlearning}, an innovative framework that allows efficient guaranteed forgetting in LLM. This framework makes unlearning easier, more cost-effective, and scalable compared to traditional exact unlearning techniques, enabling organizations to handle frequent unlearning requests.

The two core techniques of our framework, DP2U-SGD and DP2U-MLM, achieve forgetting quality and model utility nearly equivalent to retraining from scratch, while offering formal forgetting guarantees at a significantly lower cost. Although DP-SGD provides robust performance, it requires more computational resources than DP-MLM, which offers a more resource-efficient alternative without compromising the efficacy of unlearning. These methods provide a practical solution for real-time unlearning in privacy-centric applications, making \textit{DP2Unlearning} a viable choice for organizations adhering to privacy regulations such as GDPR and CCPA.

Our experiments have validated the hypotheses set forth in the methodology section that DP-aware techniques can effectively balance forgetting quality and model utility, with $\epsilon$ providing a tunable trade-off between disclosure protection and model performance. Furthermore, fine-tuning the model allows it to recover the utility lost by the enforcement of DP: with reasonably low computational cost (approximately one-fifth of RFS-R), we can achieve \emph{effective approximate unlearning}, while with moderate computational cost (approximately one-half of RFS-R), we can ensure \emph{guaranteed unlearning}.
These results demonstrate that \textit{DP2Unlearning} offers a scalable and practical solution to privacy-preserving unlearning.

As a limitation, the DP-induced disclosure protection increases the computational demands during the initial model training and requires additional storage for its permanent retention. However, this is a one-time effort that compensates for the efficiency improvements achieved at the guaranteed unlearning stage. On the other hand, our experiments rely on the TOFU benchmark datasets, which, while useful for controlled evaluation, may not fully capture the complexity and unpredictability of real-world forgetting scenarios. Future work will focus on evaluating DP2Unlearning 
with more diverse data distributions and unlearning requirements.



\textit{DP2Unlearning} can be extended to support other data types and model architectures, such as generative models for images. However, unlike textual data, the application of DP-MLM to images presents challenges due to the high dimensionality and sensitivity of pixel-level perturbations, which can severely impact image fidelity, particularly when strong guarantees (\emph{i.e.}, low $\epsilon$) are required. In such scenarios, DP-SGD may be a more viable alternative, as it operates at the model parameter level and avoids direct data perturbation, although it comes with a higher computational cost. We also aim to explore alternative privacy frameworks, such as $k$-anonymity and local DP, which may better align with specific data modalities or operational constraints. 

\section*{Acknowledgments}

Partial support for this work has been received from Notre
Dame University-IBM Technology Ethics Lab, the Government of Catalonia (ICREA Acad\`emia Prizes to J. Domingo-Ferrer and to D. S\'anchez, and grant 2021SGR-00115), MCIN/AEI/ 10.13039/501100011033 and ``ERDF A way of making Europe'' under grant PID2021-123637NB-I00 ``CURLING'', and the EU's NextGenerationEU/PRTR via INCIBE (project ``HERMES'' and INCIBE-URV cybersecurity chair).

\appendix
\section{}
This appendix contains the following sections:

{\ref{appendix:1}}: Detailed results across four baseline data sets. {\ref{appendix:2}}: Evolution of FQ and MU across epochs. {\ref{appendix:3}}: RL, TR and CP scores on four baseline data sets. {\ref{appendix:4}}: Evaluation metrics for different $\epsilon$ on Phi models. {\ref{appendix:5}}: Evaluation metrics for different $\epsilon$ on Llama2 models.

\newpage
\subsection{Detailed results across four baseline data sets}
\label{appendix:1}

\begin{table}[ht!]
    \centering       \captionsetup{labelformat=empty}
    \caption{Table \ref{tableA.8} summarizes the performance of various methods (for the Phi model), including DP2U-SGD and DP2U-MLM 
    with $\epsilon=1$. The metrics for GA, GD, PO, and KL are sourced from the TOFU~\citep{maini2024tofu} leader-board (\protect\url{https://huggingface.co/spaces/locuslab/tofu_leaderboard}). Our approaches align closely with the results from RFS-R and show a significant improvement over the baseline results reported by TOFU. 
    Best results boldfaced, second-best underlined, reference emphasized.}
   \label{tableA.8}
    \tiny
    \rotatebox{90}{
    \setlength{\tabcolsep}{3pt}
    \renewcommand{\arraystretch}{1.1}
    \begin{tabular}{l|c|ccc|ccc|ccc|ccc|c|c}
        \toprule
        \textbf{Forget} & \textbf{Method} & \multicolumn{3}{c|}{\textbf{Retain}} & \multicolumn{3}{c|}{\textbf{Real Author}} & \multicolumn{3}{c|}{\textbf{Real World}} & \multicolumn{3}{c|}{\textbf{Forget}} & \textbf{MU} & \textbf{FQ} \\
         \textbf{Ratio}& & \textbf{RL} & \textbf{TR} & \textbf{CP} & \textbf{RL} & \textbf{TR} & \textbf{CP} & \textbf{RL} & \textbf{TR} & \textbf{CP} & \textbf{RL} & \textbf{TR} & \textbf{CP} &(\%)$\uparrow$ &(\%)$\uparrow$ \\
         &&(\%)$\uparrow$&(\%)$\uparrow$&(\%)$\uparrow$&(\%)$\uparrow$&(\%)$\uparrow$&(\%)$\uparrow$&(\%)$\uparrow$&(\%)$\uparrow$&(\%)$\uparrow$&(\%)$\downarrow$&(\%)$\uparrow$&(\%)$\downarrow$\\
        \midrule
        \multirow{8}{*}{1\%} 
        & \textit{RFS-R} & \textit{99.91} & \textit{50.21} & \textit{98.73} & \textit{59.45} & \textit{43.98} & \textit{37.17} & \textit{82.41} & \textit{47.19} & \textit{39.45} & \textit{46.00} & \textit{64.39} & \textit{13.75} & \textit{54.48} & \textit{100.00}
 \\

        & GA & 75.17 & 47.84 & 84.58 & 41.23 & 46.60 & 37.99 & 75.31 & 49.08 & 41.93 & 48.35 & 52.35 & 30.18 & \underline{51.26} & 0.68 \\

        & GD & 81.01 & 47.89 & {87.42} & 35.23 & \underline{47.16} & \textbf{38.38} & 74.86 & \underline{49.65} & \textbf{42.18} &{49.74} & 51.63 & {34.86} & 50.70 & 0.30 \\

        & PO & {82.34} & 46.02 & {88.41} & 38.73 & \textbf{47.22} & \underline{38.33} & \textbf{77.71} & \textbf{50.69} & \underline{41.99} & \underline{47.74} &{52.50} & {75.23} & \textbf{51.52} & 0.13 \\

        & KL & 73.62 & 47.82 & 84.06 & {39.73} & 46.58 & 38.14 & 73.18 & 49.02 & 41.95 & 48.65 & 52.50 & 27.67 & 50.80 & 0.68 \\

        & \scalebox{0.8}{DP2U-SGD} & \textbf{99.45} & \textbf{51.03} & \underline{99.05} & \underline{42.40} & 38.55 & 33.20 & 75.60 & 45.28 & 38.35 & 47.96 & \underline{65.19} & \textbf{16.22} & 49.78 & \textbf{99.00} \\

        & \scalebox{0.8}{DP2U-MLM} & \underline{99.17} & \underline{50.24} & \textbf{99.23} & \textbf{45.80} & 39.14 & {33.35} & \underline{76.31} & 44.71 & 38.23 & \textbf{47.69} & \textbf{66.31} & \underline{16.53} & 50.26 & \textbf{99.00} \\
        \midrule

        \multirow{8}{*}{5\%} 
        & \textit{RFS-R} & \textit{99.99} & \textit{50.00} & \textit{98.74} & \textit{57.98} & \textit{45.59} & \textit{37.76} & \textit{83.69} & \textit{47.35} & \textit{39.42} & \textit{46.25} & \textit{61.59} & \textit{10.72} & \textit{54.80} & \textit{100.00}
 \\

        & GA & 27.50 & 34.64 & 3.56 & 1.37 & 38.50 & 32.86 & 18.55 & 41.30 & 39.25 & \underline{25.77} & \underline{61.98} & \textbf{1.64} & 7.27 & 14.21 \\

        & GD & 50.72 & 44.49 & 57.08 & 21.20 & \textbf{45.06} & \textbf{36.94} & 61.07 & {48.04} & \textbf{42.18} & {39.55} & 54.37 & {12.95} & 41.47 & 1.1E-03 \\

        & PO & 33.76 & 43.95 & {75.36} & 17.73 & \underline{43.11} & \underline{36.17} & {67.66} & \textbf{49.90} & \underline{41.24} & \textbf{21.86} & {52.54} & {67.57} & {38.80} & 6.9E-07 \\

        & KL & 30.32 & 35.19 & 5.56 & 1.53 & 40.29 & 34.01 & 22.75 & 42.43 & 39.83 & 28.76 & 61.20 & \underline{2.38} & 8.65 & 14.21 \\

        & \scalebox{0.8}{DP2U-SGD} & \underline{99.43} & \textbf{50.77} & \underline{99.10} & \underline{48.47} & 37.69 & 32.73 & \underline{80.94} & 44.40 & 38.37 & 44.13 & 61.77 & 12.29 & \underline{50.42} & \textbf{{98.78}} \\

        & \scalebox{0.8}{DP2U-MLM} & \textbf{99.64} & \underline{49.33} & \textbf{99.22} & \textbf{49.58} & {40.97} & {33.91} & \textbf{81.05} & \underline{48.14} & {40.40} & 46.10 & \textbf{62.76} & 14.25 & \textbf{52.23} & \underline{92.38} \\

        \midrule
        \multirow{8}{*}{10\%} 
        & \textit{RFS-R} & \textit{99.79} & \textit{49.52} & \textit{98.71} & \textit{57.20} & \textit{41.52} & \textit{35.63} & \textit{84.76} & \textit{46.27} & \textit{38.49} & \textit{45.57} & \textit{62.17} & \textit{10.46} & \textit{53.14} & \textit{100.00}
 \\

        & GA & 37.54 & 40.80 & 20.34 & 5.03 & 40.81 & 33.77 & 50.67 & 45.76 & \underline{41.64} & 36.86 & 56.98 & 15.18 & 21.49 & 0.01 \\

        & GD & 40.49 & 46.83 & 38.03 & 9.53 & \textbf{49.64} & \textbf{38.09} & 42.76 & 46.93 & 41.36 & \underline{25.06} & 56.34 & \textbf{1.19} & 30.77 & 2.65 \\

        & PO & 42.25 & 44.19 & {76.83} & 15.40 & 41.33 & \underline{35.25} & {66.52} & \textbf{49.93} & 41.16 & \textbf{22.02} & {53.09} & {72.15} & 38.11 & 2.9E-09 \\

        & KL & 38.42 & 41.46 & 23.80 & 7.37 & \underline{41.56} & 34.51 & 53.09 & 46.32 & \textbf{{41.79}} & {37.82} & 56.29 & 17.08 & 26.06 & 1.7E-03 \\

        & \scalebox{0.8}{DP2U-SGD} & \textbf{{99.64}} & \textbf{50.84} & \textbf{99.13} & \underline{45.52} & 38.54 & {33.02} & \textbf{81.23} & \underline{47.98} & 40.18 & {44.79} & \underline{62.84} & \underline{11.52} & \underline{51.13} & \underline{90.03} \\
               
        & \scalebox{0.8}{DP2U-MLM} & \underline{99.08} & \underline{49.58} & \underline{99.11} & \textbf{46.18} & {39.84} & {33.70} & \underline{80.48} & 47.89 & 39.64 & 45.98 & \textbf{63.33} & 13.38 & \textbf{51.34} & \textbf{90.14} \\
        \bottomrule
    \end{tabular}} 
\end{table}

\newpage
\begin{table}[ht!]
    \centering
    \captionsetup{labelformat=empty}
    \caption{Table \ref{tableA.9} summarizes the performance of various methods (for the Llama2 model), including DP2U-MLM with $\epsilon=1$. The metrics for GA, GD, PO, and KL are sourced from the TOFU~\cite{maini2024tofu} leader-board (\protect\url{https://huggingface.co/spaces/locuslab/tofu_leaderboard}). Our approaches align closely with the results from RFS-R and show a significant improvement over the baseline results reported by TOFU.
    Best results boldfaced, second-best underlined, reference emphasized.
}
    \label{tableA.9}
    \tiny
    \rotatebox{90}{
    \setlength{\tabcolsep}{3pt}
    \renewcommand{\arraystretch}{1.1}
    \begin{tabular}{l|c|ccc|ccc|ccc|ccc|c|c}
        \toprule
        \textbf{Forget} & \textbf{Method} & \multicolumn{3}{c|}{\textbf{Retain}} & \multicolumn{3}{c|}{\textbf{Real Author}} & \multicolumn{3}{c|}{\textbf{Real World}} & \multicolumn{3}{c|}{\textbf{Forget}} & \textbf{MU} & \textbf{FQ} \\
         \textbf{Ratio}& & \textbf{RL} & \textbf{TR} & \textbf{CP} & \textbf{RL} & \textbf{TR} & \textbf{CP} & \textbf{RL} & \textbf{TR} & \textbf{CP} & \textbf{RL} & \textbf{TR} & \textbf{CP} & (\%)$\uparrow$ & (\%)$\uparrow$ \\
         &&(\%)$\uparrow$&(\%)$\uparrow$&(\%)$\uparrow$&(\%)$\uparrow$&(\%)$\uparrow$&(\%)$\uparrow$&(\%)$\uparrow$&(\%)$\uparrow$&(\%)$\uparrow$&(\%)$\downarrow$&(\%)$\uparrow$&(\%)$\downarrow$\\
        \midrule
        \multirow{6}{*}{1\%} 
        & \textit{RFS-R} & \textit{99.8} & \textit{55.1} & \textit{99.7} & \textit{57.6} & \textit{62.9} & \textit{48.7} & \textit{69.97} & \textit{48.15} & \textit{38.09} & \textit{40.78} & \textit{67.53} & \textit{12.82} & \textit{58.70} & \textit{100.00}
 \\
        & GA & 89.9 & 47.6 & 96.9 & \underline{90.55} & \underline{57.17} & \underline{43.31} & \underline{88.03} & 53.81 & \underline{40.87} & 49.09 & 56.23 & \underline{45.99} & \textbf{60.61} & 1.43 \\
        & GD & 90.5 & \underline{47.7} & \underline{97.0} & 89.80 & 56.78 & 42.99 & 88.03 & \underline{54.14} & 40.75 & 50.33 & 55.44 & 47.82 & \underline{60.53} & 0.68 \\
        & PO & \underline{92.3} & 45.1 & \textbf{97.2} & \textbf{94.1} & \textbf{60.6} & \textbf{46.5} & 88.0 & \textbf{54.5} & \textbf{44.1} & \underline{42.3} & \textbf{88.3} & {62.4} & 1.43 & 1.43 \\
        & KL & 89.8 & 47.6 & 96.9 & 90.55 & 57.0 & 43.23 & \textbf{88.9} & 53.75 & 40.85 & 50.45 & 56.20 & 46.00 & 60.6 & \underline{1.43} \\
        & DP2U-MLM & \textbf{97.7} & \textbf{53.5} & 95.8 & {43.5} & {53.4} & 40.92 & 59.29 & 45.65 & 36.00 & \textbf{38.95} & \underline{66.8} & \textbf{19.17} & {52.3} & \textbf{99.99} \\

        \midrule
        \multirow{6}{*}{5\%} 
        & \textit{RFS-R} & \textit{99.8} & \textit{60.0} & \textit{99.7} & \textit{62.75} & \textit{64.6} & \textit{49.71} & \textit{70.54} & \textit{49.13} & \textit{40.79} & \textit{38.63} & \textit{66.80} & \textit{10.15} & \textit{58.11} & \textit{100.00}
 \\
        & GA & 0.1 & 14.9 & 1E-24 & 0.00 & 43.0 & 28.6 & 0.00 & 35.85 & 25.25 & \textbf{0.00} & \textbf{67.55} & \underline{3.3E-25} & 0.00 & 1.5E-05 \\
        & GD & 54.8 & 10.6 & 15.8 & \underline{39.95} & \textbf{57.0} & 40.68 & \textbf{75.71} & \textbf{55.07} & \textbf{42.44} & 1.38 & 64.07 & 0.02 & 30.24 & 3.1E-10 \\
        & PO & \underline{67.7} & \underline{43.1} & \underline{91.3} & 19.8 & 51.59 & \underline{40.71} & \underline{75.4} & \underline{48.7} & \underline{40.15} & 5.84 & 58.22 & 80.63 & \underline{44.53} & 2.5E-08 \\
        & KL & 22.2 & 0.0 & 2E-30 & 0.00 & 44.0 & 26.82 & 0.00 & 42.07 & 29.41 & \textbf{0.00} & 57.16 & \textbf{8.3E-31} & 0.00 & \underline{1.5E-05} \\
        & DP2U-MLM & \textbf{97.4}& \textbf{52.0} & \textbf{94.6} & \textbf{50.2} & \underline{54.5} & \textbf{41.6} & {60.4} & {45.9} & {35.3} & {40.0} & \underline{67.1} & {16.1} & \textbf{53.2} & \textbf{86.6} \\

        \midrule
        \multirow{6}{*}{10\%} 
        & \textit{RFS-R} & \textit{99.8} & \textit{55.6} & \textit{99.9} & \textit{54.52} & \textit{56.86} & \textit{43.12} & \textit{72.54} & \textit{48.15} & \textit{37.65} & \textit{37.79} & \textit{66.24} & \textit{10.16} & \textit{56.88} & \textit{100.00}
 \\
        & GA & 7.63 & 0.0 & 5E-34 & 0.00 & 36.8 & 23.21 & 0.00 & 38.30 & 23.93 & \underline{0.17} & \underline{84.71} & \textbf{7.7E-34} & 0.00 & 1.4E-20 \\
        & GD & 49.2 & \underline{46.5} & 54.7 & \textbf{80.73} & \textbf{72.99} & \textbf{56.34} & \textbf{88.9} & \textbf{60.6} & \textbf{46.37} & 0.24 & 82.20 & 1.5E-29 & \textbf{58.72} & 2.8E-23 \\
        & PO & \underline{77.1} & 45.5 & \underline{94.1} & 50.47 & 50.58 & 39.39 & \underline{84.8} & 44.23 & 36.09 & 5.65 & {54.7} & \underline{84.8} & 53.57 & \underline{1E-14} \\
        & KL & 6.7 & 0.0 & 2E-29 & 0.00 & 38.58 & 27.72 & 0.00 & 39.51 & 25.95 & \textbf{0.00} & \textbf{86.6} & \underline{1E-30} & 0.00 & 5E-27 \\
        & DP2U-MLM & \textbf{97.4} &\textbf{53.6} &  \textbf{{94.2}} & \underline{54.9} & \underline{52.0} & \underline{39.8} & {62.1} & \underline{46.1} & \underline{36.5} & {40.6} & {67.8} & \underline{16.1} & \underline{53.8} & \textbf{17.6} \\
        \bottomrule
    \end{tabular}} 
\end{table}

\newpage
\subsection{Evolution of FQ and MU across epochs}
\label{appendix:2}

\begin{figure}[ht]
    \centering
    \begin{minipage}{1\textwidth}
        \centering        \includegraphics[width=0.9\textwidth]{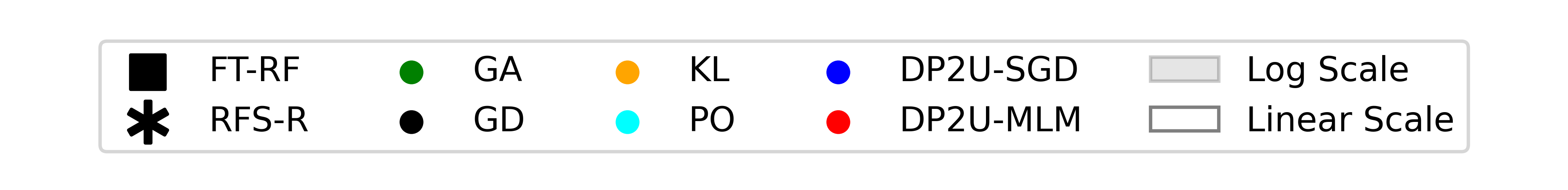}
    \end{minipage}
    \begin{minipage}{0.47\textwidth}
        \centering    \includegraphics[width=1\textwidth]{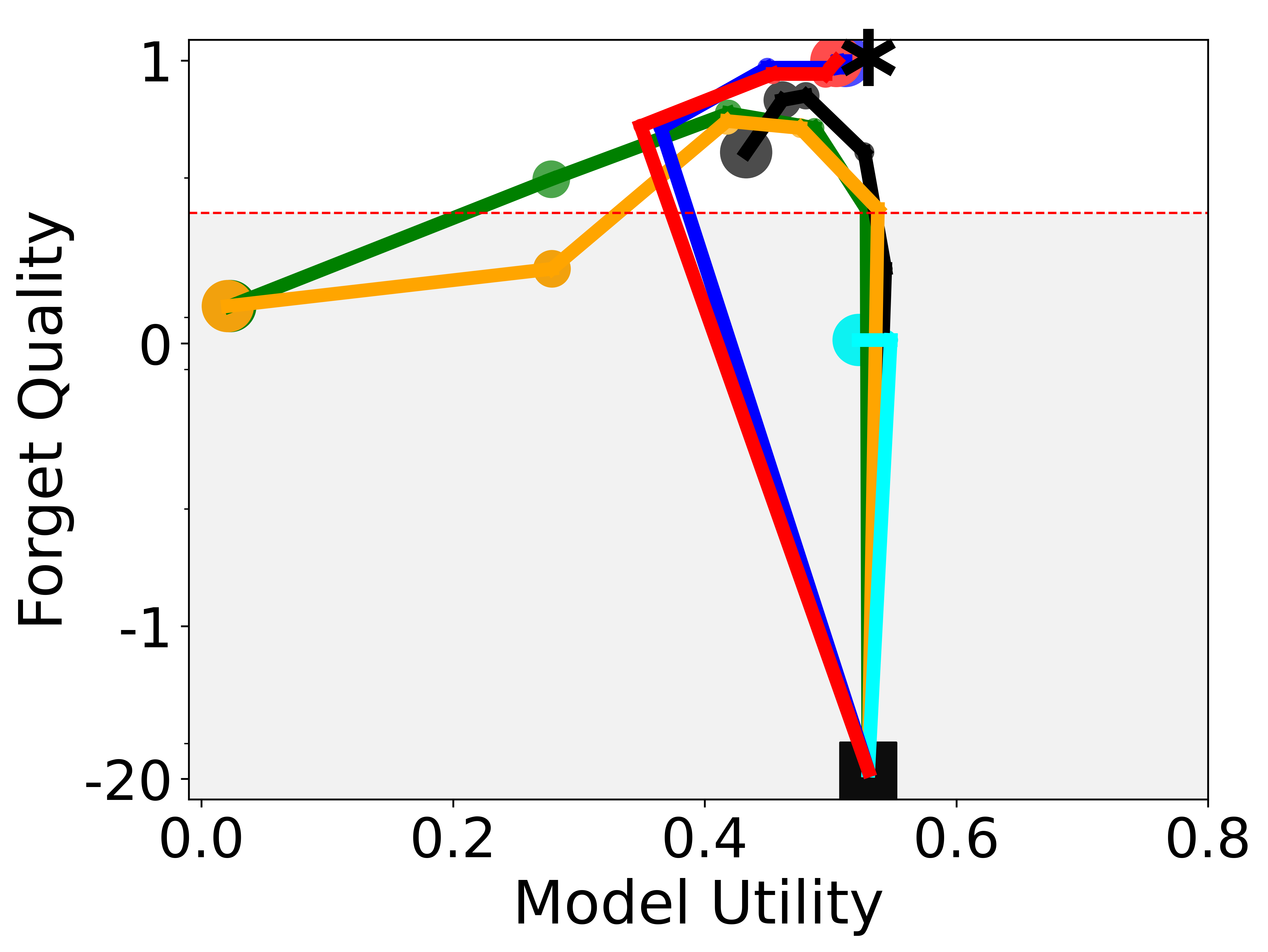}
    \end{minipage} 
    \begin{minipage}{0.43\textwidth}
        \centering        \includegraphics[width=1\textwidth]{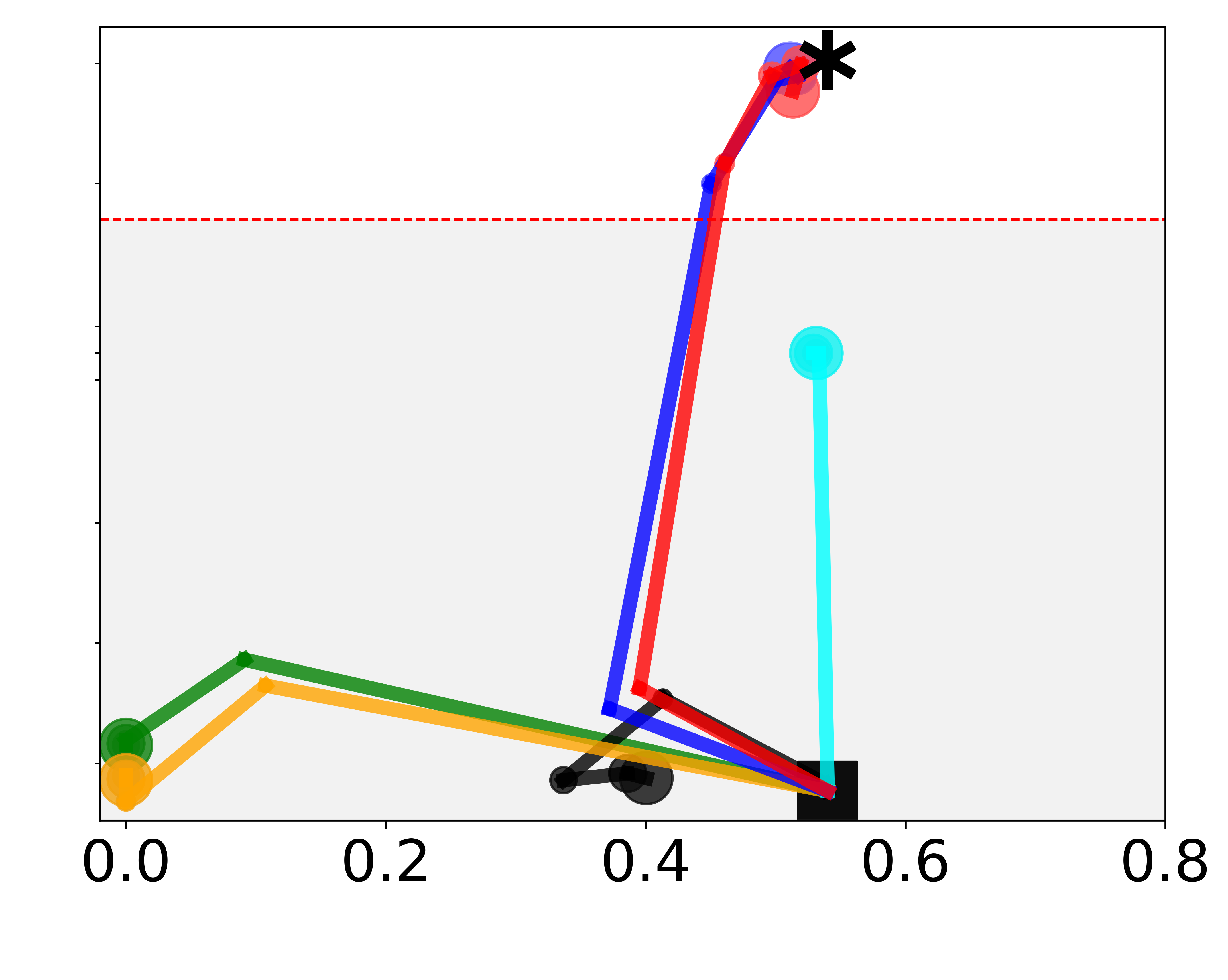}      \end{minipage}
    
    \vspace{0.3cm}    
    \begin{minipage}{0.47\textwidth}
        \centering    \includegraphics[width=1\textwidth]{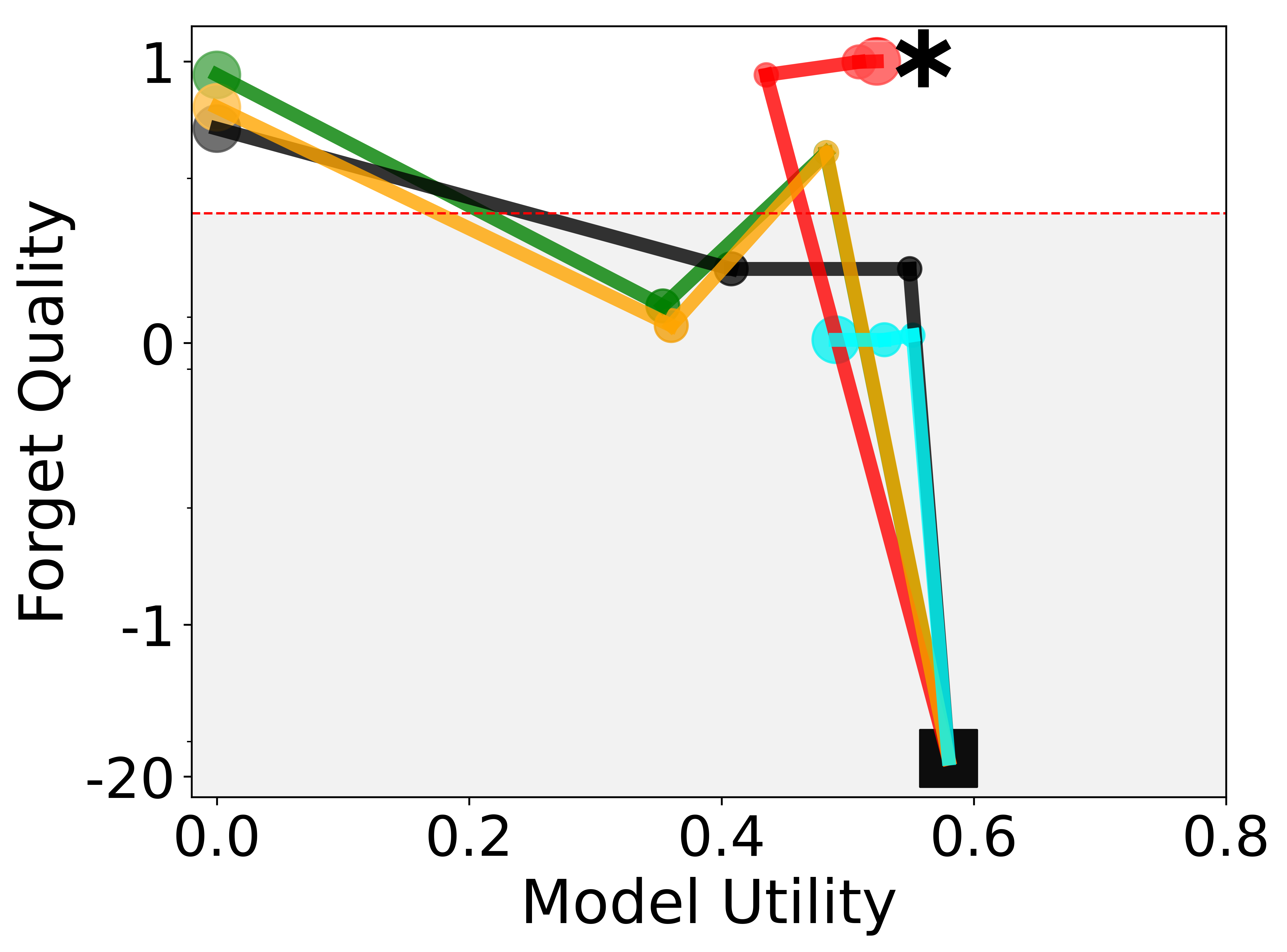}
    \end{minipage} 
    \hspace{0cm}
    \begin{minipage}{0.43\textwidth}
        \centering        \includegraphics[width=1\textwidth]{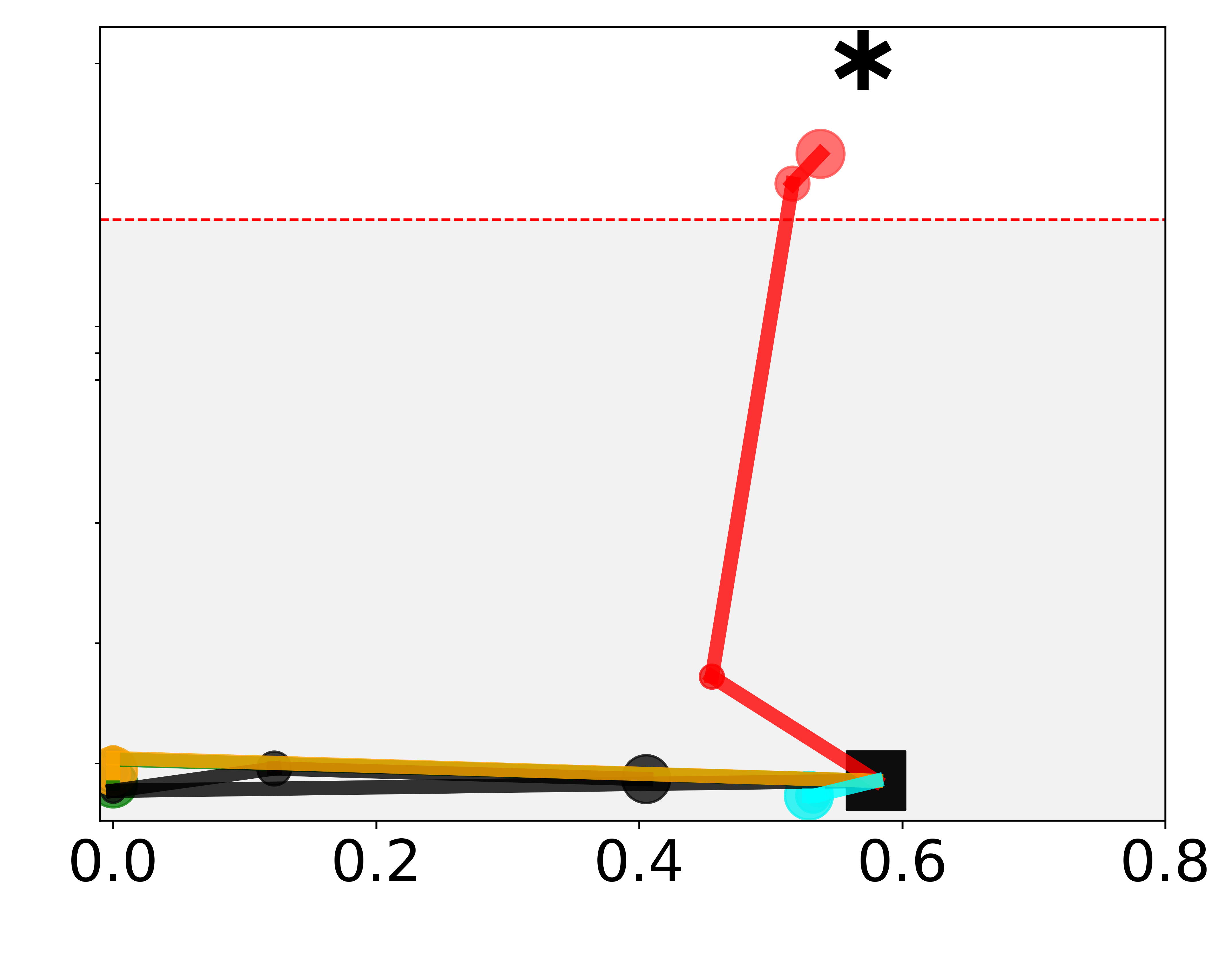}      \end{minipage}
    \caption{Forget quality vs. model utility for the Phi (Row 1) and Llama2 (Row 2) models across different unlearning methods, forget ratios and epochs. The forget ratios are set to 1\% (left) and 10\% (right) of the total training data. $E=10$ for FT-RF and RFS-R, while $E'=5$ for other methods.}
    \label{figureA.9}
\end{figure}

\newpage
\subsection{RL, TR and CP scores on four baseline data sets}
\label{appendix:3}

\begin{figure}[ht!]
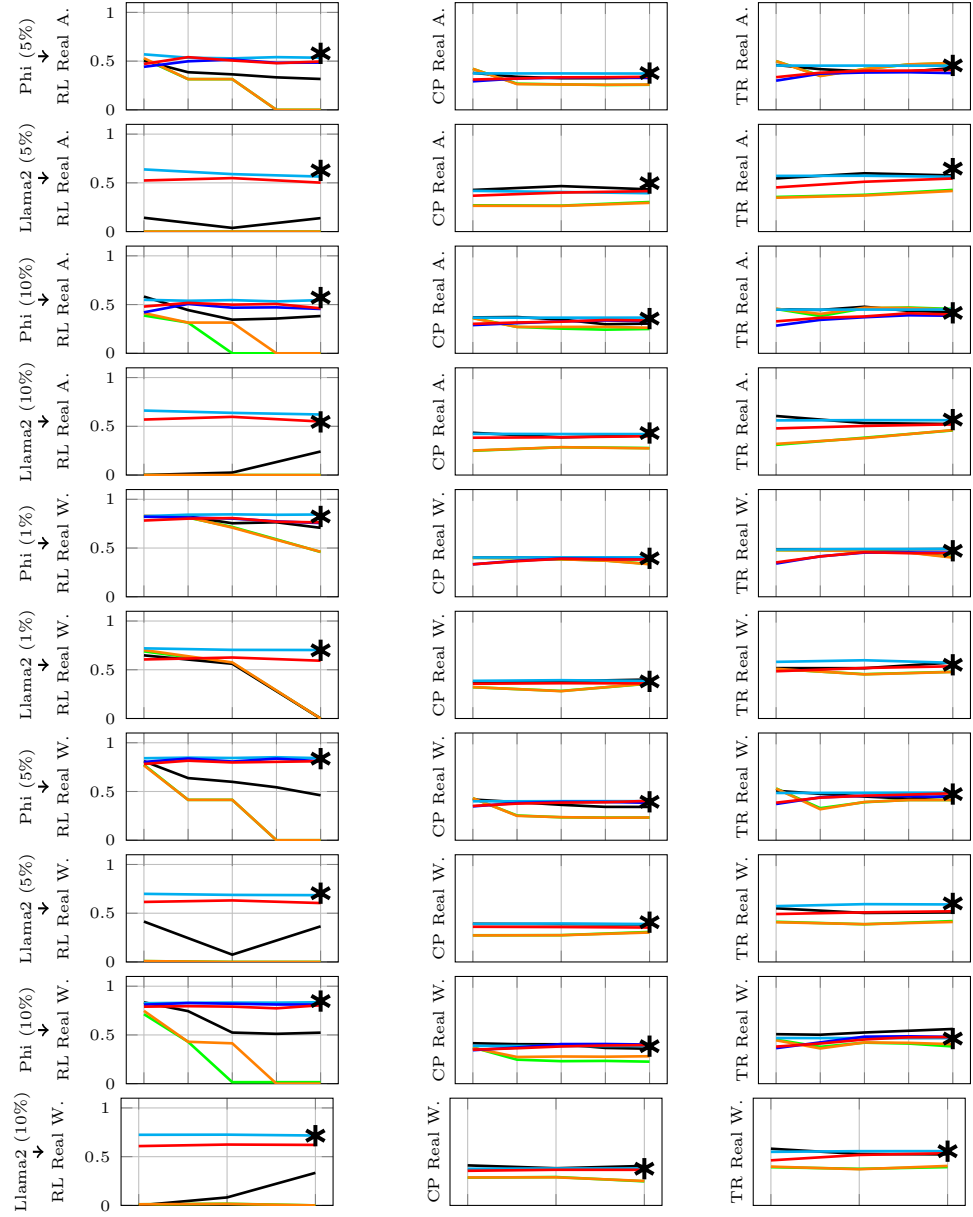

    \centering

    \end{minipage}
    \vspace{-0.9cm}
    \caption{Evaluation metrics span across methods, epochs, and forget ratios. For RFS-R, $E=10$ (Phi) and $E=6$ (Llama2).}
\end{figure}

\newpage
\subsection{Evaluation metrics for different $\epsilon$ on Phi models}
\label{appendix:4}

\begin{figure}[ht!]
    \centering
    %
\end{minipage}

\vspace{-0.3cm}
    \caption{Evaluation metrics across privacy budgets. $E=10$ (RFS-R) and $E'=5$ (for others). For each metric, forget ratios: 1\% (left), 5\% (middle), and 10\% (right).}
\end{figure}

\newpage
\subsection{Evaluation metrics for different $\epsilon$ on Llama2 models}
\label{appendix:5}

\begin{figure}[ht!]
    \centering
    %
\end{minipage}

\vspace{-0.2cm}
    \caption{Evaluation metrics across privacy budgets. $E=6$ (RFS-R) and $E'=3$ (for others). For each metric, the forget ratios are 1\% (left), 5\% (middle), and 10\% (right).}
\end{figure}
\bibliographystyle{elsarticle-harv} 
\bibliography{elsarticle}

\end{document}